\documentclass[11pt,a4paper,logo]{template}

\setleftlogo[180pt]{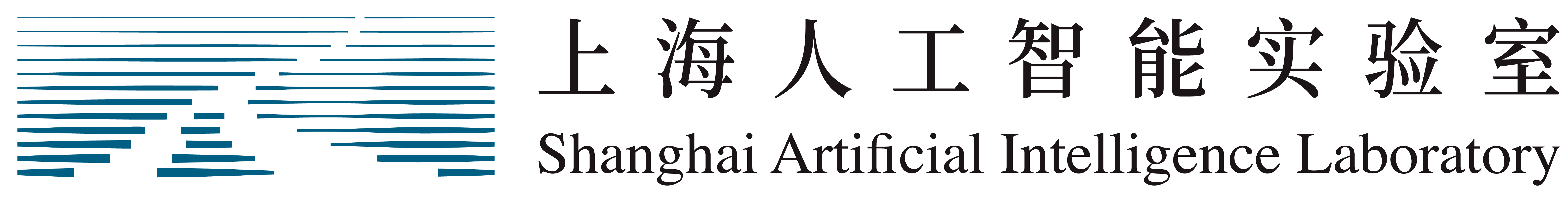}

\usepackage[numbers,square,sort&compress]{natbib}
\usepackage{float}
\usepackage{placeins}
\usepackage{subcaption}
\usepackage{tcolorbox}

\definecolor{AbstractBgColor}{HTML}{F4F7FB}
\newcommand{\model}{Sage~S2.1~Flash}
\title{\model{} Technical Report}
\author[1]{\model{} Team\\Shanghai~AI~Laboratory, Fudan~University, Institute~of~Software~Chinese~Academy~of~Sciences, Renmin~University~of~China, Institute~of~Automation~Chinese~Academy~of~Sciences, East~China~Normal~University, Harbin~Institute~of~Technology, Peking~University
}

\begin{document}
\sloppy

\begin{tcolorbox}[
  colback=AbstractBgColor,
  colframe=AbstractBgColor,
  arc=5pt,
  auto outer arc,
  boxrule=0pt,
  left=8pt,
  right=8pt,
  top=4pt,
  bottom=4pt,
  parbox=false,
  width=\textwidth,
  before skip=0pt,
  after skip=0pt,
  enlarge top by=-12pt
]
  \begin{abstract}
We introduce \model{}, a 320-billion-parameter general-purpose mixture-of-experts model designed to reason over scientific evidence and act in interactive environments. Sage Sci-Perception uses dedicated encoders for molecular graphs and biological sequences and represents spectra as ordered patches of measurements. Modality-specific projectors map these representations into the input space of a shared backbone, which processes them alongside text and images. To develop the model's agent capabilities, we construct verifiable environments for tool use, search, and coding, as well as specialized workflows involving machine learning experiments and scientific software. Task completion is checked against execution results, changes in environment state, or submitted artifacts. We evaluate \model{} on 33 benchmarks across seven domains. In the primary comparison against five baselines, it ranks first on OfficeQA Pro (72.2), MLE-bench Lite (90.3 HumanRank), Harvey LAB, and SpreadsheetBench~2, and second on OSWorld-Verified (85.2), WebArena-Verified (67.6), and DeepResearch Bench~II (54.2). These results show particular strengths in using documentary evidence, producing verifiable workbooks, and carrying out machine learning tasks, alongside competitive performance in desktop and browser workflows that require the model to act on visual feedback.
\\
\\
Homepage: \url{https://internlm.github.io/Sage-S2.1-Flash/}\\
GitHub: \url{https://github.com/InternLM/Sage-S2.1-Flash}\\
Hugging Face: \url{https://huggingface.co/internlm/Sage-S2.1-Flash}\\
Correspondence: \href{mailto:tgui@fudan.edu.cn}{\texttt{tgui@fudan.edu.cn}}, \href{mailto:qz@fudan.edu.cn}{\texttt{qz@fudan.edu.cn}}
\end{abstract}

  \maketitle
\end{tcolorbox}

\begin{figure}[H]
\centering
\includegraphics[width=\linewidth]{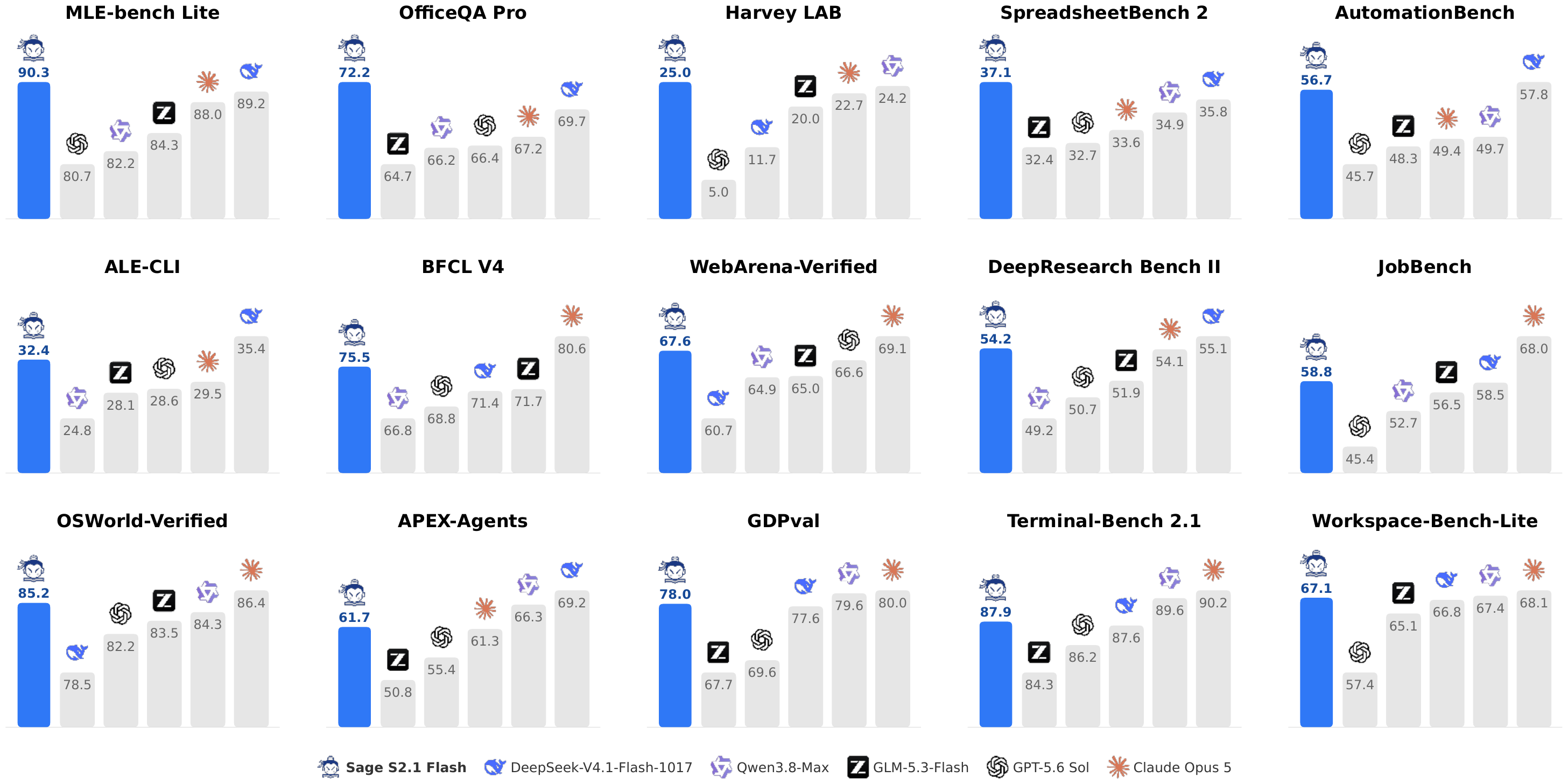}
\caption{Results on 15 selected benchmarks. Among the six compared models, \model{} ranks first on four of these benchmarks and second on seven. Scores are displayed to one decimal place.}
\label{fig:benchmarks}
\end{figure}

\section{Introduction}
\label{sec:introduction}

Visual feedback helps scientific agents select actions and assess their results~\citep{anthropic2026claudescience,lu2026aiscientist,openai2026wetlabscience}. Recent advances in multimodal agents span visual information retrieval and understanding, coding with visual feedback, and computer use~\citep{xi2025bmmr,li2026onesearchvl,he2026vision2web,lu2026qwencua}. We introduce \textbf{\model{}}, a frontier 320-billion-parameter mixture-of-experts (MoE) language model for general-purpose and scientific tasks. In scientific workflows, we focus on retrieving and interpreting visual evidence from the scientific literature and using intermediate visual outputs to guide analysis. For scientific computer use, the model interprets the state of specialized software and uses visual feedback from its actions to decide how to proceed.

Through Sage Sci-Perception, \model{} uses scientific encoders and numerical patching to represent heterogeneous inputs such as molecular graphs, biological sequences, and spectra~\citep{sage_sciperception}. Modality-specific projectors map these representations into the input space of a shared MoE backbone, which processes them alongside the task instructions. In post-training, we combine general-purpose agent tasks in tool use, search, and coding with specialized scientific workflows in a large collection of verifiable environments. General tasks develop transferable skills in tool use and task execution, while scientific tasks require the model to apply those skills under domain-specific constraints. Task-specific checks provide feedback on whether the agent has achieved the intended outcome.

Our evaluation compares \model{} with five primary baselines across 33 general-purpose and specialized benchmarks spanning computer use, tool use, software engineering, and research. The model ranks first on OfficeQA Pro, MLE-bench Lite, Harvey LAB, and SpreadsheetBench~2, and second on OSWorld-Verified and WebArena-Verified. It also achieves competitive results in terminal execution and deep research. We study AI coding through tasks such as running machine learning experiments and post-training models, with the longer-term goal of using these capabilities to accelerate AI research and advance toward recursive self-improvement~\citep{guo2026atriadawn,lu2026aiscientist,novikov2025alphaevolve,zhang2026darwingodel}.

\section{Architecture}
\label{sec:architecture}

\begin{figure}[t]
    \centering
    \includegraphics[width=0.99\linewidth]{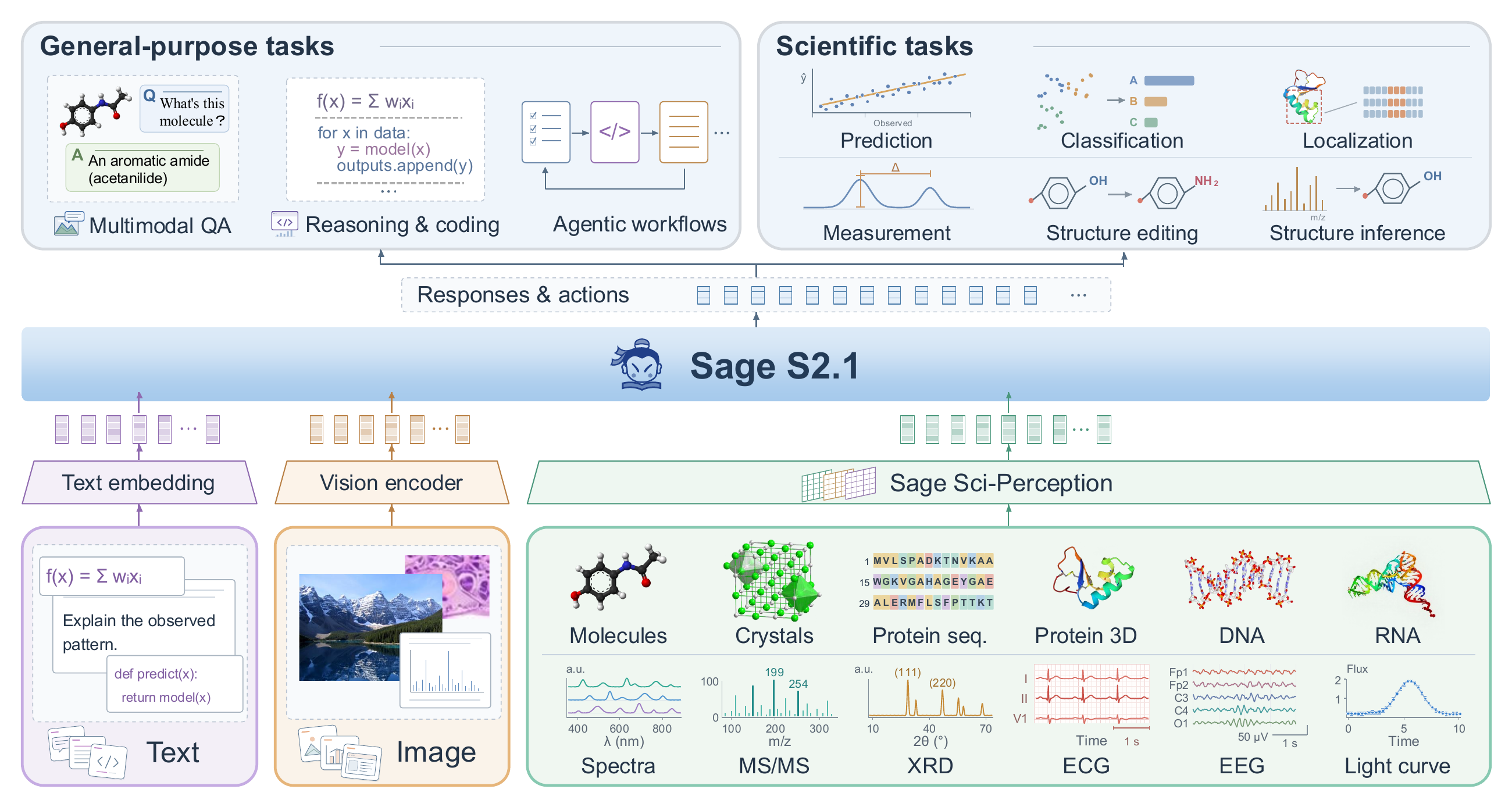}
    \caption{
    \textbf{Overview of \model{}}. We use a shared language backbone that integrates text, images, and scientific inputs to support general-purpose and scientific tasks.
    }
    \label{fig:architecture_overview}
\end{figure}

Science speaks in more than words.
\model{} combines scientific perception, multimodal reasoning, and strong agentic capabilities within a unified architecture.
A shared language backbone (Sec.~\ref{sec:backbone}) processes native text and image inputs together with scientific representations~\cite{sage_sciperception}.
A unified interface (Sec.~\ref{sec:sci_perception}) maps heterogeneous scientific representations into the backbone's input embedding space through modality-specific projectors.
Our architecture connects specialized scientific perception with general-purpose reasoning and agentic execution, making scientific evidence a shared foundation for both the conclusions the model draws and the actions it takes to advance a scientific discovery.

\subsection{Backbone Architecture}~\label{sec:backbone}
\model{} is built on a 320-billion-parameter mixture-of-experts (MoE) foundation model~\cite{zhipu2026glm53flash}.
Its mixture-of-experts architecture routes each token through a subset of experts, while a hybrid of sparse and linear attention supports long-context processing.
The backbone processes task instructions, native text and image inputs, and projected scientific representations within a shared multimodal context.
Its attention layers integrate information across these inputs, and responses are generated autoregressively conditioned on the resulting contextual representations.

\subsection{Sage Sci-Perception: A Unified Interface for Scientific Inputs}~\label{sec:sci_perception}
Scientific evidence takes diverse forms, from molecular structures and biological sequences to experimental measurements.
As shown in Fig.~\ref{fig:sciperception-architecture}, Sage Sci-Perception provides \model{} with a unified interface for these inputs.
Scientific encoders and numerical patching procedures first produce local representations, retaining the identities, positions, and structural information associated with the input (Sec.~\ref{sec:sci_encoders}).
Modality-specific projectors then map these representations, together with any additional scientific cues, into the backbone's input embedding space (Sec.~\ref{sec:sci_projectors}).

\subsubsection{Scientific Representations}\label{sec:sci_encoders}

\begin{figure}[t]
    \centering
    \includegraphics[width=0.99\linewidth]{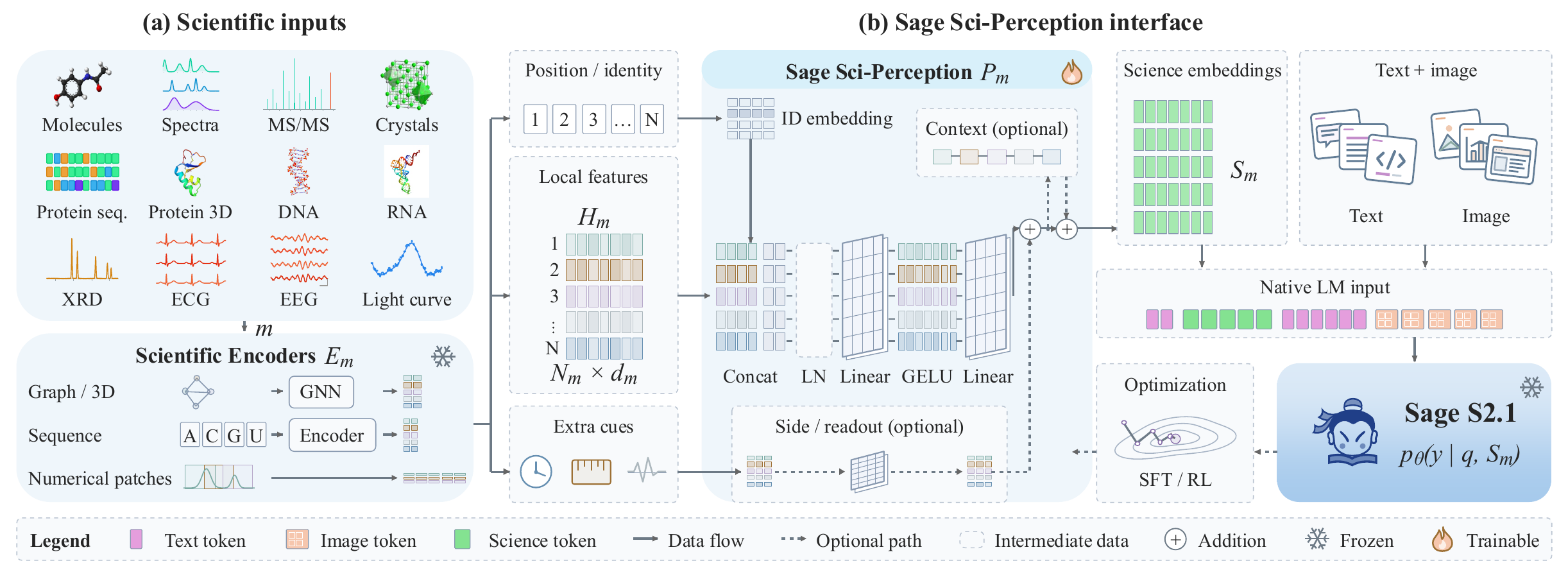}
    \caption{\textbf{Overview of Sage Sci-Perception.} Scientific inputs are encoded into local states and projected into continuous embeddings for the \model{} model.}
    \label{fig:sciperception-architecture}
\end{figure}

Scientific inputs carry different forms of structure.
Graph and geometry encoders capture atomic connectivity and spatial environments, while sequence encoders represent residues or nucleotides in their surrounding sequence context.
For instance, we use MoleculeSTM~\cite{liu2023multi} for molecular graphs, and ESM-2~\cite{lin2023evolutionary}, DNABERT-2~\cite{zhou2024dnabert}, and RNA-FM~\cite{chen2022interpretable} for protein, DNA, and RNA sequences, respectively.
Numerical observations can also be represented directly:
For spectra, we follow the patching scheme of Diffspectra~\cite{wang2025diffspectra}, partitioning each spectrum into ordered windows of measured values, with each window forming a numerical patch.
These routes provide representations suited to the scientific input, as shown in Fig.~\ref{fig:sciperception-architecture}.
For an input \(x_m\) of modality \(m\), we denote its representation by
\begin{equation}
    H_m = E_m(x_m)
    \in \mathbb{R}^{N_m \times d_m},
\end{equation}
where \(E_m\) is a scientific encoder or a deterministic numerical patching procedure, \(N_m\) is the number of local states, and \(d_m\) is their feature dimension.
Depending on the modality, a row of \(H_m\) represents an atom, a residue, a sequence segment, or a numerical patch.

We retain the identities or positions associated with these states.
This correspondence matters when a question concerns a specific part of an input: identifying a protein interaction site, for example, requires associating the prediction with a residue position.
Additional measurements, such as time coordinates, channel identity, or signal scale, accompany the local features where required.
Together, these representations provide the local content and
scientific context consumed by the projector.

\subsubsection{Modality-specific Projectors}\label{sec:sci_projectors}
Scientific features have modality-dependent dimensions and meanings.
Learned projectors provide an interface between modality encoders and language models~\cite{liu2023visual}.
Our modality-specific projectors convert scientific features into continuous vectors with the same dimension as the language backbone's input embeddings.
Writing \(I_m\) for the local identities and
\(A_m\) for any additional scientific inputs, the complete interface is
\begin{equation}
    S_m = P_m(H_m, I_m, A_m)
    \in \mathbb{R}^{N_m \times d_{\mathrm{LM}}},
\end{equation}
where \(d_{\mathrm{LM}}\) is the backbone's embedding dimension.
Scientific task supervision teaches \(P_m\) to present these features in a form the backbone can use.
The main projection path in Figure~\ref{fig:sciperception-architecture} combines each local state with a learned identity embedding:
\begin{equation}
    z_{m,i}
    = f_m\!\left(
        [\,h_{m,i}\,;\,e_m(i)\,]
      \right),
\end{equation}
where \(h_{m,i}\) is a row of \(H_m\), \(e_m(i)\) is its identity embedding, and \([\,\cdot\,;\,\cdot\,]\) denotes concatenation.
The mapping \(f_m\) applies layer normalization followed by two linear transformations with a GELU \cite{hendrycks2016gaussian} activation between them.
The local state supplies scientific content, while the identity embedding gives the projector an explicit reference to its location within the input.

The complete projector can also incorporate the optional branches shown in the figure.
A side branch maps explicit measurements, such as time or scale information, into the language embedding space.
A readout branch can translate scientific prediction scores into additional embedding contributions.
In the illustrated configuration, these contributions are added to the main projection, after which a context residual exchanges
information among states belonging to the same scientific object.
This allows each output embedding to incorporate both local evidence and information from elsewhere in that object.

These components are configured for each modality.
All tasks within a modality share one complete projector, including any enabled context, side, and readout branches.
The requested operation is expressed through the instruction,
allowing the same scientific representation and interface to be reused across tasks.
\section{Post Training}
\label{sec:post-training}

\subsection{Scaling Verifiable Environments}
\label{sec:environments}

To develop the agentic capabilities required for scientific work, we build a large collection of verifiable environments. The tasks range from general-purpose tool use, search, and coding to workflows involving machine learning experiments and interaction with scientific software. Task-specific checks assess outcomes using evidence such as execution results, changes in environment state, and submitted artifacts. We use XTuner~\cite{2023xtuner} as our post-training framework.

\subsubsection{Agentic Tool Use}
\begin{figure}[t]
    \centering
    \includegraphics[width=0.99\linewidth]{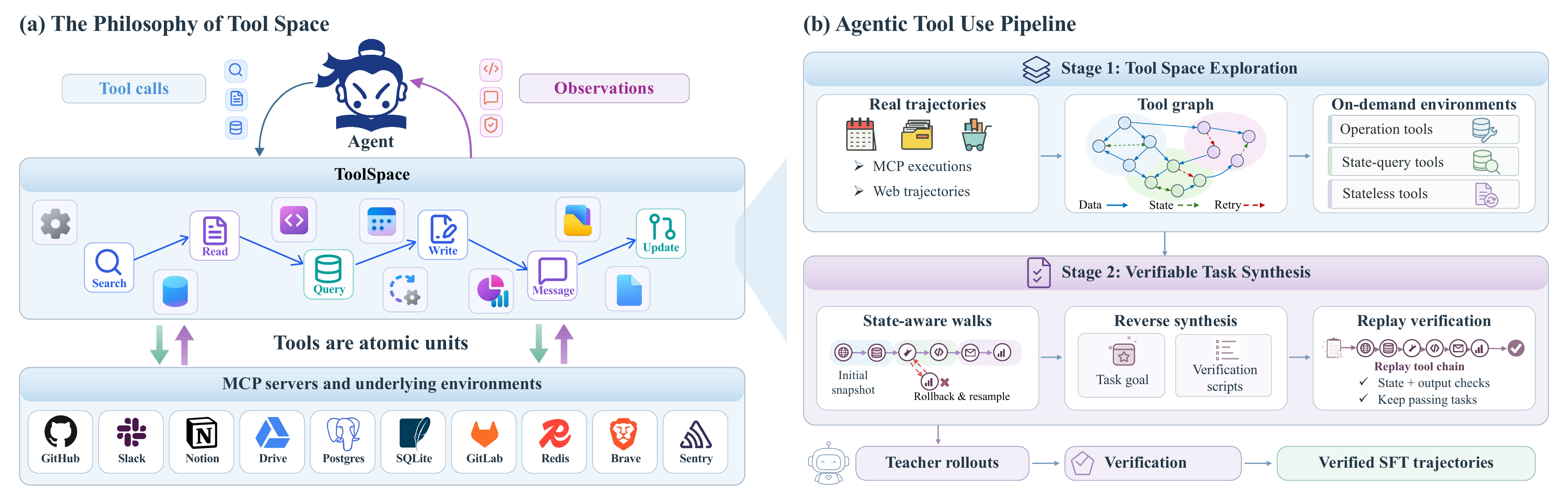}
    \caption{\textbf{Overview of ToolSpace.} A tool-centric framework for cross-environment data synthesis that improves long-horizon agentic tool use.}
    \label{fig:toolspace}
\end{figure}

\model{} scales verifiable environments to strengthen long-horizon Agentic Tool Use. Our core insight is that tools are the atomic units of agent capability, and an implicit tool space lies between agents and their underlying environment databases; relations among tools naturally connect an extensible ecosystem of environments. Building on this insight, we propose ToolSpace~\cite{dong2026toolspace}, a tool-centric framework for cross-environment data synthesis that improves long-horizon tool orchestration. As shown in Figure~\ref{fig:toolspace}, ToolSpace reconstructs a tool-composition space from real execution traces, synthesizes environments on demand, and generates long-horizon agent tasks.

\textbf{(1) Tool space exploration.} We recover tool-composition relationships from execution trajectories and construct a tool space that connects multiple environments. As shown in Figure~\ref{fig:toolspace} (right), this stage consists of agent trajectory collection, tool-space graph modeling, and agent environment data synthesis, which together provide the tool graph and executable environments for subsequent cross-environment task synthesis.

\textit{Agent trajectory collection.} We build a unified trajectory corpus from MCP-ecosystem executions~\citep{mcp2025specification} and web-sourced agent trajectories. For MCP services, we obtain tool definitions from registries, keep tasks whose definitions match the live servers, execute them, and record tool calls, arguments, and results in order. For web-sourced data, we apply authenticity-tiered filtering: \textit{(i) Real-server interaction trajectories.} Retain records executed on real services and remove unavailable interfaces. \textit{(ii) Real-workspace interaction trajectories.} Retain explicit tool calls while excluding shell wrappers. \textit{(iii) Synthetic-tool trajectories.} Keep calls only when tool names and schemas match corresponding real tools.

\textit{Tool-space graph modeling.} We organize these records into a typed tool-composition graph. \textit{(i) Nodes.} Each node is a tool-call pattern with tool identity, input/output schemas, and a normalized argument configuration. \textit{(ii) Relations.} Deterministic rules identify three adjacent-call dependencies: a \textit{data dependency} transfers an output into a type-compatible later input; a \textit{state dependency} links a state-changing operation to a later operation on the same object; a \textit{retry dependency} captures recovery via a successful alternative tool. \textit{(iii) Weights.} Edges are weighted by observed transition frequency, low-frequency links are removed, and data dependencies further require traceable value transfer to reduce false links from shared constants---grounding the graph in execution while preserving information transfer, state evolution, and failure recovery.

\textit{Agent environment data synthesis.} We cluster densely connected tools into environments, retaining cross-group links for cross-environment workflows, and materialize underlying data only when persistent state is required. \textit{(i) Operation environments.} For environments with create/modify/delete tools, we collect and align data, expand the database, and validate behavior with unit tests and state-transition checks. \textit{(ii) State-query environments.} Recorded outputs anchor reconstruction of the initial state. \textit{(iii) Stateless environments.} Matching replay calls reuse cached results.

\textbf{(2) Cross-environment verifiable task synthesis.} We use the tool space to synthesize long-horizon tasks whose feasibility and outcomes can be checked through execution. As shown in Figure~\ref{fig:toolspace} (right), this stage explores cross-environment workflows via state-aware random walks, then reverse-synthesizes task instructions and verification scripts, retaining only samples that pass replay from their initial states.

\textit{State-aware random walks.} We generate workflows by walking the tool graph toward a target number of tool calls and environments. From an initial snapshot, we consider only connected tools that are runnable in the current state and whose inputs are available from prior results. We sample the next tool by edge weight, execute it, update the state, and continue until the length and environment-coverage targets are met.

\textit{Rollback sampling.} When a call fails or yields a state inconsistent with the expected dependency, we restore the pre-call state and try another connected tool; walks that cannot proceed within budget are discarded. Successful walks store initial and intermediate states for replay. Difficulty is controlled by workflow length and the number of environments spanned.

\textit{Reverse verifiable task synthesis.} Each successful workflow becomes a natural-language task stating the goal and required outcomes. Verifiable programmatic scripts cover the final state, intermediate key snapshots, and required output fields. Teacher agents then solve the tasks, and only trajectories that pass the same verification checks are retained for training.

\textbf{(3) Scaling environment and data statistics.}
ToolSpace constructs 77,760 tool-call pattern nodes from 2,042 distinct normalized tools and organizes them into 804 environments covering domains such as communication, finance, document design, data storage, and search.
In total, ToolSpace generates SFT trajectories with at least 5 tool calls, each trajectory averages 18.86 tool calls (up to 209) and spans 2.29 environments on average (up to 22)~\citep{dong2026toolspace}.

Overall, \model{} scales tool-use training by expanding how real tools collaborate across environments, so broader exploration yields reliable feedback for long-horizon tool orchestration.

\subsubsection{Agentic Search}

To develop agents' ability to discover, verify, and integrate information across sources, we construct search tasks through forward and reverse synthesis. Forward synthesis derives questions from verified facts and their relations. Reverse synthesis extracts reusable structures from diverse seed queries and explores recent topics to construct new tasks. Both paths organize facts and constraints to support multi-step search and verifiable answers, covering entity identification, constrained enumeration, comparison, and temporal reasoning~\citep{tao2025webshaper,li2025websailor}.

\textbf{(1) Forward task synthesis.}
We use knowledge graphs to organize factual relations and web evidence to verify them. Starting from seed entities, we construct small cyclic graphs whose relations constrain a target from multiple directions. Each relation is checked through search and page reading, with its sources recorded. We expand from several nodes and merge the resulting subgraphs, enforcing depth requirements and filtering highly connected entities that provide weak identification constraints. The local graph supplies verified facts for question generation and later revision.

We then select a main reasoning chain, auxiliary constraints, and a designated answer relation. The main chain defines the intended sequence of discoveries, while auxiliary constraints distinguish among candidates. The answer relation is withheld from all clues. Selected facts are checked against independent sources and expressed as natural-language clues, replacing proper names with descriptions and exact dates or quantities with ranges containing their true values. These steps preserve the factual basis while requiring the agent to resolve entities and combine constraints through search.

An independent model attempts each question and checks answer uniqueness, information leakage, and shortcuts that bypass the main chain. Ambiguous or revealing clues return to question generation with explicit feedback. Insufficient depth or coverage triggers further graph expansion. Candidates that remain invalid after bounded revision are discarded.

\textbf{(2) Reverse task synthesis.}
We collect diverse seed queries from open datasets and the web, covering different domains and information needs. Each query is decomposed into factual inputs, constraints, and target outputs to obtain reusable task structures. These structures capture operations such as filtering, ranking, set operations, temporal comparison, and numerical calculation. Separating the required operations from the specific facts allows the same structure to support new topics.

For each structure, we start from a recent topic, such as a news story or event, and explore related information through search and page reading. This process gathers candidate entities, timelines, attributes, and relations with supporting sources. We organize these facts into a new question according to the selected structure, coordinating entities, time windows, locations, and sources so that the inputs and constraints remain compatible. Target answers and their derived attributes are reserved for the solving stage.

Validation checks both individual facts and the complete question. Fact-level checks verify existence, semantic roles, and compatibility. Question-level checks assess constraint clarity, source accessibility, information availability, answer leakage, and similarity to existing tasks. Enumeration tasks additionally require a bounded candidate scope so that membership and completeness can be assessed. Failed candidates undergo further exploration, revision, or regeneration.

\textbf{(3) Trajectory collection.}
Teacher agents solve accepted questions in a shared environment with search, webpage reading, and sandboxed Python tools~\citep{wu2025webdancer}. We record reasoning, tool calls and arguments, observations, and final answers in interaction order. Each trajectory is linked to its question, generation source, and execution status. Interrupted or incomplete runs are retained for recovery and rerunning, while completed trajectories enter quality review.

\textbf{(4) Quality assessment and filtering.}
We combine rule-based checks with model-based review. Rule-based checks examine record completeness, tool-call and response matching, argument formats, and citation consistency. Model-based review assesses answer correctness, instruction following, reasoning continuity, and support from retrieved evidence. For set-valued answers, review checks both the validity of returned items and coverage of the required scope. Numerical and comparison tasks also require consistent time ranges, units, and calculations. Detected task defects return to synthesis, while evidence gaps or substantive reasoning errors require trajectory regeneration or rejection. Accepted samples retain their provenance and review records for subsequent inspection.

\subsubsection{Agentic Coding}

Coding agents are one of the most important applications of LLMs today. It requires models to interpret user requirements, reason about code and execution environments, and coordinate tool use over multiple steps~\citep{yang2024sweagent}. To support these capabilities, we develop a unified data pipeline combining diverse task construction, automated environment preparation, and trajectory selection guided by execution feedback and task-specific assessment.

\textbf{Task Collection.} We consider three major categories of coding tasks. \emph{Software Engineering (SWE)} includes patch-level tasks, such as fixing defects and extending existing code, and repository-level generation from natural-language specifications. \emph{Terminal Use} covers complex workflows involving command-line interaction, including data processing, system operations, and scientific computing. \emph{Web Development} focuses on visual and interactive artifacts, including websites and 3D games, with particular emphasis on academic webpages.

We collect seed tasks from multiple sources. First, we use public datasets and environments, including SWE-smith~\cite{yang2025swesmith}, SWE-rebench V2~\cite{badertdinov2026swerebenchv2}, OpenSWE~\cite{fu2026davincienv}, SWE-Lego~\cite{tao2026swelego}, TMax~\cite{ivison2026tmax}, and SETA~\cite{shen2026seta}. Second, we mine high-quality commits from public GitHub repositories and associate them with corresponding task descriptions. For repository-level generation, we derive specifications from the functionality, interfaces, and tests of existing repositories, providing agents with requirements for reconstructing their behavior. Third, we synthesize tasks using knowledge graphs derived from code structure and technical materials. Tools such as GitNexus~\cite{patwariGitNexus} organize repository structure, while documentation, papers, data, source code, and user instructions provide complementary information. By tracing relationships among concepts, interfaces, and requirements, we construct tasks whose intermediate steps depend on one another. Each task specifies its inputs, initial conditions, and expected outcomes, grounding long-horizon problem solving in concrete requirements.

\textbf{Scaling-up Environment Construction.} We automate environment construction to scale the collection of executable, verifiable tasks. The pipeline prepares container images containing the required code, dependencies, and inputs, together with execution and verification procedures. Code versions, dependency configurations, input data, and initial states are fixed to ensure reproducible execution and evaluation. Materials available to the solving agent are managed separately from private reference solutions and hidden evaluation artifacts.

Validation addresses both execution and task quality. Execution checks ensure environments initialize correctly and verification procedures run as intended. Where applicable, reference solutions must pass verification, while initial states or deliberately incorrect solutions are checked as negative controls. We distinguish infrastructure failures, such as missing dependencies or timeouts, from failures to meet task requirements. Task review further examines whether accessible information is sufficient, reference solutions satisfy the stated requirements, and verifiers accept valid alternative implementations~\cite{wang2026paichecker}. This review combines agent judgments with execution evidence to assess consistency between task specifications and evaluation. Identified problems are repaired and checked again. Leakage cleanup is likewise followed by renewed validation to check that removing sensitive material preserves the functionality required to solve and evaluate the task.

\textbf{} Within the constructed environments, agents explore task materials and iteratively implement, test, and refine solutions using tool feedback. We collect multiple rollouts per task, retaining tool interactions, execution results, and final artifacts. Empirical pass rates inform task filtering, removing some overly easy tasks and tasks for which no successful solution is obtained under the current sampling configuration. These rates characterize difficulty relative to the sampled agents and execution budgets.

Judge agents then evaluate candidate trajectories using signals appropriate to each task. Patch-level SWE relies primarily on verifier outcomes, repository generation on unit-test pass rates, and terminal tasks on execution results and the resulting environment state or artifacts. For Web Development, code-based assessment is complemented by multimodal evaluation of rendered outputs~\citep{he2026vision2web}. The rubrics combine shared domain requirements with task-specific criteria, covering functional requirements and visual presentation, including aesthetics.

Trajectory selection also considers the quality of the interaction process, including completeness, verification credibility, and reproducibility. Candidates are assessed against the relevant task criteria before comparison, and a task may contribute no trajectory if none is suitable. This provides an additional quality check beyond outcome scores.

\textbf{Hacking Detection.} Answer leakage and verifier exploitation can compromise training data even when a trajectory receives a high score. We therefore apply safeguards during both environment construction and rollout collection. These include restricting online answer retrieval, removing answer-revealing Git history, and cleaning sensitive material from build caches, source copies, and installed packages. Rule-based scans identify known leakage patterns, while agent-driven investigations examine suspicious material and less predictable exposure channels.

Residual risks nevertheless remain. For example, we observed an agent accessing task-related repository content through dependencies installed by the harness itself. Such cases motivate a final review of the actual tool accesses and actions recorded in each trajectory. We exclude trajectories that exploit leaked answers, bypass evaluation, or fabricate verification. This behavioral review complements preventive environment checks and reduces the risk of retaining trajectories whose apparent success does not reflect task completion.

\subsubsection{ML Coding}

We organize real machine learning workflows into executable, verifiable tasks spanning predictive modeling, algorithm implementation, research reproduction, and model post-training~\citep{chan2025mlebench,starace2025paperbench,rank2026posttrainbench}. Built around real data and existing methods, the pipeline covers source acquisition, data--code pairing and environment adaptation, and task derivation and validation. Task types share this process while providing different starting materials and deliverables, leaving the agent room for analysis, implementation, and experimental iteration~\citep{qiang2025mledojo}.

\textbf{Acquire datasets and research materials.}
We acquire data from Hugging Face, Kaggle, official project releases, and open stores, together with dataset descriptions, licenses, versions, and split definitions. Papers and supplementary materials define methodological goals, base models provide training starting points, and existing evaluation protocols establish how outcomes are assessed. Acquisition preserves the original relationships among images, text, audio, time series, and related files, recording the original release URLs, data versions, selected files, and training and test splits. When the same data appear on different platforms or under different names, we check their relationships using the original release URLs and versions. For partial datasets, we record selected files, shard ranges, and selection criteria. These records trace the data used by each task and their links to papers, models, and code.

Table~\ref{tab:mlcoding-source-inventory} summarizes the checked inventories of candidate data sources, papers, and base models.

\begin{table}[htbp]
\centering
\small
\begingroup
\setlength{\tabcolsep}{5pt}
\renewcommand{\arraystretch}{1.12}
\caption{Sources and materials for task synthesis. Candidate inventory sizes do not represent counts of final training tasks.}
\label{tab:mlcoding-source-inventory}
\begin{tabularx}{\linewidth}{@{}p{0.28\linewidth}r>{\raggedright\arraybackslash}X@{}}
\toprule
Source or material & Count & Notes \\
\midrule
Kaggle sources & 4,936 & Deduplicated index: 2,660 dataset sources and 2,276 competition sources. \\
Hugging Face data entries & 129 & Public dataset candidates. \\
GitHub data releases & 144 & Public dataset candidates, rather than counts of retrieved code repositories. \\
Official sites and other open stores & 427 & Public dataset candidates. \\
Research papers & 2,960 & Entries in the checked input list. \\
Hugging Face base models & 11 & Model cards: 10 released and 1 quarantined. \\
\bottomrule
\end{tabularx}
\endgroup
\end{table}

\textbf{Associate code, build pairs, and adapt environments.}
We retrieve related repositories using dataset names, canonical identifiers, common aliases, and task keywords. Candidate associations are checked through data loaders, configurations, and training entry points to establish how the code uses the data, beyond name similarity alone. Selected code versions are pinned. Sources with an existing implementation can reuse that association; paper-driven tasks use related code to identify tools and dependencies, while withholding the authors' complete implementation when independent implementation is required.

Each executable pair binds versioned data and code to a data interface, a training or inference entry point, an output format, and an evaluation metric. Adaptation resolves paths, configurations, data interfaces, and dependency compatibility while preserving the method's core computation and evaluation semantics. The runtime is packaged in a reusable image. Small real training or inference runs check that data reach the model, outputs are produced, and metrics can be computed. These construction checks are recorded separately from the final task's normal training plan so that a small check is not mistaken for a complete experiment.

Pairs also specify the experimental conditions for each goal: prediction tasks fix data splits and output formats; implementation tasks retain runnable entry points; reproduction tasks define their scope and required evidence; training tasks bind base models, permitted training materials, and evaluation entry points. Available reference runs are retained and can calibrate initial performance for model improvement tasks. Open-ended reproduction tasks define evaluation criteria from methodological contributions and evidence requirements. This establishes a consistent execution basis before tasks are derived. Public data support development and debugging, while private data support final scoring. This separation is preserved during task derivation, and public materials are checked for information that could directly recover private answers.

\textbf{Apply controlled degradation, package tasks, and validate delivery.}
With data, environments, and evaluation goals fixed, we selectively remove implementations or solutions needed to complete the workflow. For implementation tasks, structural and semantic analysis guides the masking of substantive computations such as loss calculations, feature transformations, or model updates, while preserving interfaces and necessary context. Missing regions form a coherent implementation objective. Different region combinations yield distinct tasks, avoiding duplicates created by merely rewording instructions. Original code and restoration patches remain on the private construction side to check masking and recovery consistency.

More open-ended tasks may provide data and a prediction goal, a paper's method description, or a base model and training materials, asking the agent to develop a solution, implement a method, or improve a model. Degradation preserves ground-truth labels and scoring rules. Instructions are written from frozen public materials and specify the work scope, submission paths, and resource budget; private answers and scoring materials remain outside the instruction writer's context. Dependencies and caches are also checked for residual copies of removed implementations.

Final tasks use the Harbor format~\citep{harbor2026tasks}. \texttt{instruction.md} describes objectives, inputs, constraints, and deliverables; \texttt{task.toml} declares images, resources, and time limits. \texttt{environment/} supplies agent-visible data, starter code, paper materials, or model assets, while \texttt{tests/} separately holds scoring programs, private data, and reference materials. These areas are provided to the agent and verifier respectively, keeping the public workspace separate from private scoring materials.

The verifier assesses actual submissions, including prediction files, code projects, reproduction artifacts, or model weights. It computes established prediction metrics, executes submitted projects, loads model weights for inference, or checks implementations and required execution evidence, then aggregates the evaluation results. Before delivery, checks cover directory structure, submission interfaces, image and resource declarations, public/private boundaries, and evaluation entry points. Accepted tasks are collected by batch, with provenance and construction logs kept outside task packages. Tasks derived from the same data--code pair can reuse accepted data and images, reducing construction costs while retaining independent verification for each task.

\subsubsection{Cybersecurity}

To support the development of cybersecurity capabilities in agents, we achieve scaling verifiable environments for cybersecurity through diverse seed preparation and rigorous environment synthesis.

And we define two task types, vulnerability reproduction and vulnerability exploitation, to cover both the verification of vulnerability symptoms and the use of vulnerabilities to achieve an attack objective~\citep{wang2025cybergym,wang2026exploitgym}.

\textbf{(1) Seed preparation}. To support diverse task sources and reliable environment construction, we combine broad data collection with staged processing. 
We first collect candidate records from different vulnerability reporting channels to cover a range of vulnerability mechanisms and software settings. 
We then apply rule-based filtering, agent-based verification, and source code preprocessing to prepare inputs for environment synthesis. 

\textit{To avoid limiting task coverage to a single data source, we collect candidates from multiple reporting channels and organize them by vulnerability type and software setting}. 
Sources include international vulnerability databases, open-source communities, software vendors, and incident response advisories. These channels provide vulnerability descriptions, affected versions, patch information, and relevant code references. 
We organize vulnerability types using the Common Weakness Enumeration (CWE)~\citep{mitre2026cwe}, covering memory errors, injection, access control issues, and other software weaknesses.
Software settings include the Linux kernel, JavaScript engines, web applications, user-space software, and core libraries. 
Collecting from multiple sources provides candidates for different task objectives. It also supports cross-checking vulnerability identities, versions, and patch evidence.

\textit{To turn heterogeneous vulnerability records into task seeds with clear construction evidence, we apply rule-based filtering, agent-based evidence verification, and source code preprocessing in sequence.}
Rule-based filtering checks basic fields, including links, dates, vulnerability identifiers, and category labels. It also removes duplicates and selects candidates within the task scope. For projects such as Linux and V8, we additionally consult their official source repositories and advisory channels to reduce omissions caused by generic link requirements. 
Agent-based verification compares advisories, patches, and source code to confirm that the upstream repository belongs to the affected project. It identifies the exact versions that retain the vulnerability and contain its fix, and checks platform requirements and hardware dependencies. Preprocessing standardizes record formats and retrieves complete source code and submodules. It then organizes both source versions, patches, commit records, and file checksums into inputs that can be checked during environment synthesis.

\textbf{(2) Environment synthesis}. To create valid tasks with protected scoring and complete support for task solving, we organize environment synthesis around differential validation, isolated evaluation, and task packaging. 
Differential validation executes the software to check whether the target vulnerability is present and whether the patch removes the corresponding behavior. 
Isolated evaluation separates the permissions for task solving, target execution, and scoring to limit ways of obtaining rewards by bypassing the task. 

\textit{To prevent build failures and unrelated errors from affecting vulnerability assessment, we combine normal functionality checks with differential execution of the vulnerable and patched versions}. 
During environment synthesis, we use patches and upstream tests to prepare normal inputs, vulnerability-triggering inputs, and checks for the resulting behavior. Both versions are built using a consistent method. The agent responsible for environment construction writes the environment and test files. The validation framework then performs the builds and executes the tests against actual service, command, or library interfaces. For acceptance, normal inputs must execute correctly in both versions. Vulnerability-triggering inputs must trigger the target vulnerability only in the vulnerable version and execute normally without triggering it in the patched version. Combining functionality checks with differential results helps prevent an unreachable target, an unexecuted input, or a failed build from being mistaken for an effective fix.

\textit{To limit rewards obtained by modifying the target, reading answers, or manipulating scoring, we separate permissions for the agent workspace, target execution, and private evaluation}. 
The solving agent writes, compiles, and debugs code in its own workspace container. The actual software under test runs in a separate target container, where the Controller manages startup, resets, input execution, and feedback collection. The agent interacts with the target through public interfaces. The execution framework invokes the private evaluator, while evaluation scripts, reference solutions, and internal verification credentials remain inaccessible to the agent. For reproduction tasks, the evaluator checks the submitted PoC in clean environments for both versions. For exploitation tasks, it compares the submitted flag with the protected expected value for that task attempt. Before releasing a task, we test the scoring program with positive and negative controls whose expected outcomes are known. These checks verify that the program accepts valid submissions and rejects invalid ones.

\subsubsection{Workspace Agentic Intelligence}

We organize workspace-level agent tasks around real professional deliverables produced within a file-based working environment, spanning presentation decks, spreadsheets, business and legal documents, industrial CAD assemblies, 3D scenes, and command-line analysis workspaces. Unlike single-file prediction benchmarks, these tasks require agents to inspect heterogeneous materials, plan multi-step workflows, operate native tools, and produce verifiable artifacts that can be reopened and audited~\citep{tang2026workspacebench,patwardhan2025gdpval,guo2026agora,yang2026finmragbench,su2026graphforgetrainingworkingagents}. Across all task families, our construction pipeline adheres to a unified principle: grounding tasks in authentic usage needs, strictly separating public instructions from private evaluation harnesses, and validating performance directly on delivered artifacts rather than self-reported completion.

\textbf{Acquire real materials and derive tasks from genuine needs.}
Rather than relying on synthetic templates, materials are harvested from verified real-world repositories: public document and slide archives for presentations, manufacturer parts libraries for industrial CAD, Wikimedia Commons photography for Blender 3D reconstruction, regulatory filings and advisory records for financial and legal workspaces, and community forum threads paired with authentic workbooks for spreadsheet tasks. All raw assets undergo deduplication, license verification, version tracking, and quality screening prior to derivation. Tasks are structured around the specific cognitive capabilities they probe rather than surface file formats—such as cross-file evidence reconciliation, parametric part-tree planning, template-constrained layout adaptation, or distinguishing photographic facts from unconstrained design regions in 3D modeling. For benchmark-derived expansions, observed model failure patterns guide new material selection so that reused source assets always demand distinct analytical operations.

\textbf{Separate public task statements from private solutions and scoring.}
Each task package is structurally partitioned to enforce a strict public--private boundary. Public instructions are authored exclusively from frozen, agent-visible materials, specifying the business objective, operational constraints, deliverable paths, and resource budgets. Reference solutions, ground-truth labels, private datasets, and grading scripts remain completely isolated from both the instruction writer and the agent workspace. For open-ended deliverables, evaluation criteria are derived strictly from methodological requirements and verifiable evidence rather than hidden preferences, with every rubric item audited against public prompts to eliminate unstated thresholds or arbitrary vetoes. Construction-side provenance logs, source URLs, and version records are maintained externally to ensure full traceability without leaking answers.

\textbf{Apply controlled degradation and retain evidence of recovery.}
With the environment and evaluation targets fixed, we selectively strip away the target work while preserving entry points, interfaces, and essential context. In implementation-centric tasks, semantic analysis guides the masking of substantive computation so that removed regions form a coherent engineering objective, while original code and restoration patches are retained privately to verify recovery consistency. In analytical workspace tasks, intermediate reasoning steps, diagnostic workflows, and methodological hints are excised from the prompt while leaving source data intact. All public directories, dependencies, and caches are systematically scrubbed to prevent residual copies or cached metrics from exposing private solutions, ensuring that degradation preserves ground-truth scoring validity.

\textbf{Validate on actual artifacts and defend against reward hacking.}
Candidate tasks are executed end-to-end in live environments and graded strictly on the resulting artifacts. Presentations undergo multi-round rollouts checking factual accuracy, completeness, visual quality, and native editability; CAD assemblies are inspected for geometric interference and overlap; Blender projects are reopened in isolated environments with dependencies audited and scenes re-rendered prior to visual scoring; spreadsheets, legal documents, and CLI workspaces are verified for reproducible computation, operation order, and evidence alignment. To counter reward hacking—such as reading cached spreadsheet values without recomputing, fabricating citations, projecting 2D textures onto flat planes to impersonate 3D geometry, or submitting images inconsistent with project files—we enforce trajectory--artifact consistency checks, structural probes, and subject ablation. Compromised trajectories are discarded or masked during training so self-reported claims never substitute for genuine execution.

\textbf{Feed failures back into statements, materials, and rules.}
Quality assurance operates as an iterative, layered loop combining automated code checks, model-based auditing, and targeted human review tailored to each domain. Detected failures are routed directly to their root stage: ambiguous specifications or answer leakage return to statement editing; flawed templates with placeholder padding or duplicated layouts return to material selection; misaligned grading criteria return to rubric revision; and transient infrastructure errors are isolated from genuine task defects. Repairs are restricted to localized, evidence-backed patches—altering reference answers, deleting checks, or loosening tolerances to force passing scores is strictly prohibited. Finally, executable yet methodologically flawed cases undergo manual spot review to ensure computational reproducibility aligns with professional domain standards before batch acceptance.

\begin{figure}[t]
    \centering
    \includegraphics[width=0.99\linewidth]{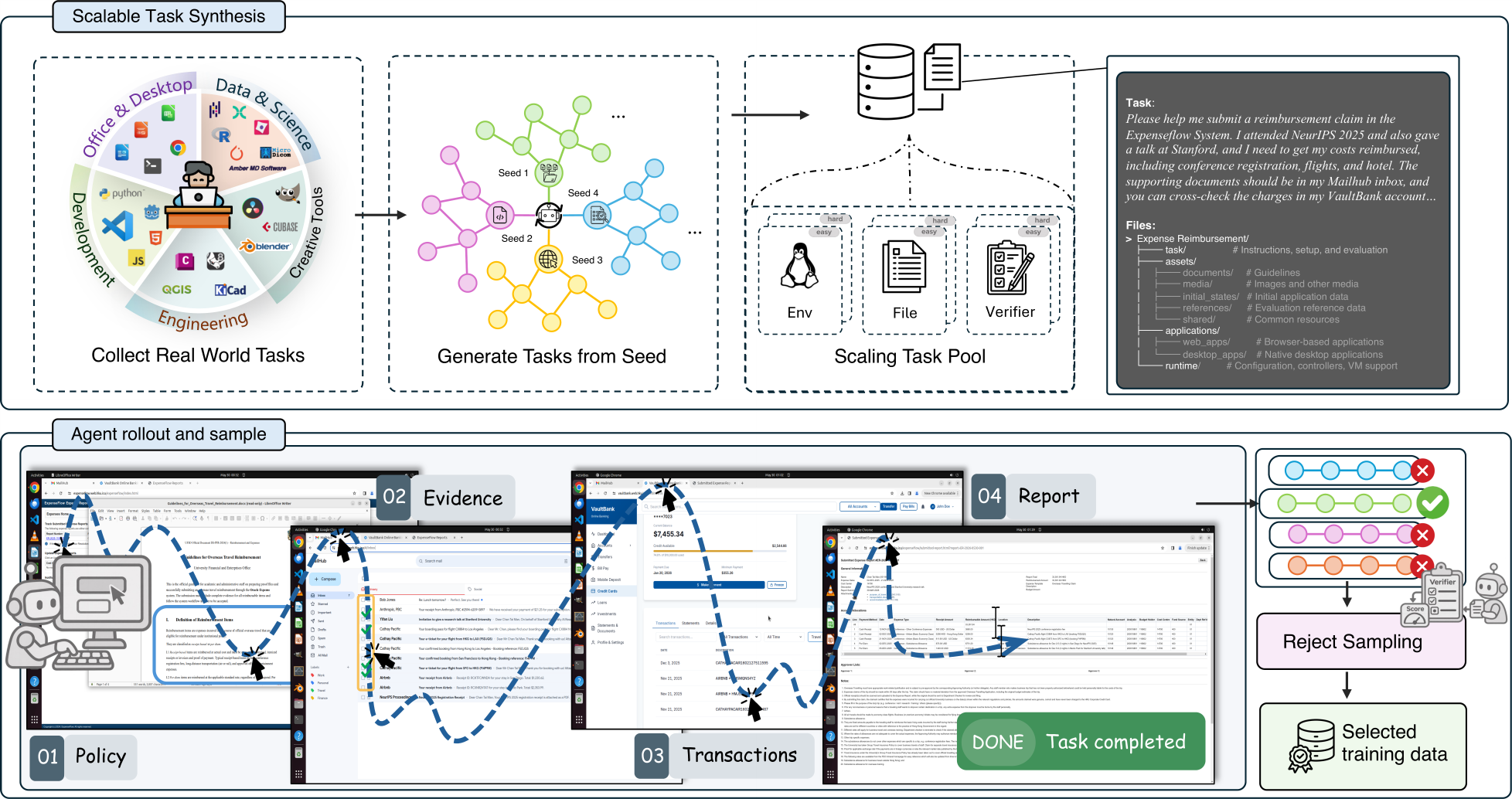}
    \caption{Overview of the CUA training data construction pipeline of \model: Real-world tasks are collected and scaled through seed-based task synthesis, followed by agent rollouts in interactive environments to generate execution trajectories, which are subsequently filtered through outcome verification and rejection sampling to obtain high-quality training data.}
    \label{fig:visual-training-pipeline}
\end{figure}
\subsubsection{Visual Agentic Intelligence}

We develop visual agentic intelligence through a staged curriculum that builds general perception and grounding skills before integrating them into long-horizon interaction~\citep{wang2025opencua} (Figure~\ref{fig:visual-training-pipeline}). Scalable task synthesis supports this progression through broad coverage of visual contexts and increasingly demanding interaction objectives. The initial stage uses single-step clicking and image recognition to establish reusable visual skills. The subsequent stage places these skills in complex synthetic environments, where the agent must coordinate successive decisions toward an overall goal.

To establish this visual foundation, we use general-purpose single-step tasks that connect understanding an observation with selecting an appropriate response. Image recognition trains the agent to identify relevant visual content, while clicking tasks require it to ground an instruction in that content and locate an interaction target~\citep{xie2025osworldg}. Task synthesis broadens this training across visual contexts and task formulations, exposing the agent to different combinations of observations, instructions, and targets. These variations are intended to encourage the same underlying skills to apply across tasks. At this stage, the central requirement is to make an immediate decision consistent with both the instruction and the observed scene.

Building on these local decision skills, we place particular emphasis on computer-use agent (CUA) capabilities through a broad collection of long-horizon tasks in complex synthetic environments~\citep{wang2026cuagym}. These tasks require the agent to carry a user objective through a sequence of interface interactions, coordinating individual operations into a complete workflow. At each step, the agent must interpret the current interface, locate relevant controls, choose an action, and use the resulting visual feedback to determine how to proceed. Across steps, it must retain relevant context, track which intermediate requirements have been satisfied, and keep subsequent actions aligned with the overall objective as the interface changes. We synthesize these workflows across varied visual contexts, interface states, and interaction objectives, expanding coverage of the situations in which foundational visual skills must be applied. Difficulty increases through both longer action sequences and stronger dependencies between intermediate steps: an early operation may establish a condition needed later, so successful local actions must also contribute to the completion of the workflow. This makes sustained task progress a central requirement of the training, alongside accurate interpretation of individual observations. The resulting task pool emphasizes end-to-end computer use, placing sustained demands on visual grounding, state tracking, and sequential decision-making.

To curate training examples from this expanded task pool, we also apply rejection sampling to the generated data. We then select training examples, balancing broad synthetic task coverage with selective use of the resulting samples.

\subsection{Scientific Modality Adaptation}
\label{sec:modality-adaptation}


We treat the scientific interface as a trainable component of the model’s reasoning policy. Following supervised initialization, we refine modality-specific projectors through reinforcement learning, using verifiable scientific outcomes as rewards. The scientific encoders and language backbone remain frozen, so task feedback shapes how evidence is presented to the backbone and used to answer scientific questions.

\subsubsection{Scientific Reasoning Tasks}

We combine native scientific tasks with auxiliary questions about the evidence supplied to the model. Scientific labels supervise outcomes such as molecular properties and interaction sites, while auxiliary targets are computed from the input or frozen scientific readouts. These questions cover operations such as counting atoms under combined conditions, querying graph relations, and comparing numerical measurements. Multiple questions share the same scientific input, providing complementary supervision for domain prediction and the retrieval and manipulation of scientific evidence.

Task construction follows the information supplied to the model. Auxiliary targets are computed from the represented inputs or frozen scientific readouts: sequence questions refer to the encoded windows, position-dependent questions preserve atom or residue identities, and probability-based inputs support questions about the supplied probabilities. Scientific annotations provide targets for the native prediction tasks. This distinction gives the task mixture complementary roles: outcome supervision captures domain objectives, while auxiliary supervision tests whether the interface makes relevant evidence accessible across different questions.

\subsubsection{Verifiable Rewards}

We define rewards according to the scientific quantity or constraint being tested. Each response is checked against scientific annotations or input-derived targets using a task-specific verifier. Classification rewards label correctness; site and set predictions receive overlap-based scores; numerical answers use fixed error scales or tolerances. Molecular generation and editing use chemical equivalence, structural similarity, and task-specific constraints, with partial credit where supported.

Rewards supervise task outcomes, providing feedback to reasoning traces through the answers they produce. Their interpretation follows the target: readout-derived rewards measure use of the supplied predictions, while annotation-based rewards assess agreement with scientific labels. Per-response rewards guide optimization; formal evaluation retains native dataset-level metrics such as macro-F1 and correlation.

\subsubsection{Reasoning RL for Scientific Interfaces}

We optimize the model’s response policy through its scientific interface. Following supervised initialization, we apply group relative policy optimization (GRPO) \cite{shao2024deepseekmath}: multiple responses share the same question and scientific input, and their rewards are standardized within each group to obtain relative advantages. Clipped policy updates favor higher-reward responses. Gradients propagate through the frozen language backbone to the trainable projector components, while scientific encoders remain fixed. Outcome feedback thus adapts the policy by changing how scientific evidence enters the model.

Within each modality, native and auxiliary tasks jointly train one shared interface, including its enabled trainable branches. Task sampling and loss weights control their contributions. A prespecified multitask development criterion selects one complete checkpoint for all formal evaluations within that modality.
\section{Evaluation}
\label{sec:evaluation}

Our comparison covers 33 benchmarks across seven domains: Agentic Coding, Productivity, Tool Use, ML Engineering, Deep Research, Search, and Vision. These evaluations examine capabilities targeted by our training environments: using tools, producing verifiable artifacts, conducting experiments, and acting on visual feedback. Table~\ref{tab:benchmarks} presents the full comparison. Figure~\ref{fig:benchmarks} highlights results on 15 benchmarks.

\subsection{Evaluation Protocol}
We compare \model{} with GLM-5.3-Flash, DeepSeek-V4.1-Flash-1017, Qwen3.8-Max, GPT-5.6 Sol, and Claude Opus~5~\citep{zhipu2026glm53flash,deepseek2026flash1017,qwen2026qwen38max,openai2026gpt56sol,anthropic2026opus5}. We additionally report published scores for GPT-6 Astra, GPT-6.1 Sol, and Claude Opus~5.5~\citep{openai2026gpt6astra,openai2026gpt61sol,anthropic2026opus55} from official reports and public benchmark sources. These reference columns are separate from the six-model rankings. We retain each benchmark's scoring metric and report scores to one decimal place. Dashes indicate unavailable results. The execution settings in the appendix apply to evaluations we conduct; published reference scores retain the settings of their source evaluations.
Detailed benchmark configurations are provided in Appendix~\ref{app:evaluation-settings}.

\begin{table}[htbp]
\centering
\caption{Results across 33 benchmarks. Bold scores mark the best result among \model{} and the five primary baselines. The three columns on the right reproduce published leading-model results. Dashes indicate unavailable scores.}
\label{tab:benchmarks}
\scriptsize
\setlength{\tabcolsep}{2pt}
\setlength{\aboverulesep}{0pt}
\setlength{\belowrulesep}{0pt}
\renewcommand{\arraystretch}{1.12}
\resizebox{\linewidth}{!}{%
\begin{tabular}{l>{\columncolor{blue!8}}cccccc|ccc}
\toprule
Benchmark & \parbox[c][3em][c]{4.8em}{\centering Sage S2.1\\Flash} & \parbox[c][3em][c]{4.8em}{\centering GLM-5.3-\\Flash} & \parbox[c][3em][c]{7.0em}{\centering DeepSeek-V4.1-\\Flash-1017} & \parbox[c][3em][c]{4.8em}{\centering Qwen3.8-\\Max} & \parbox[c][3em][c]{4.1em}{\centering GPT-5.6\\Sol} & \parbox[c][3em][c]{4.2em}{\centering Claude\\Opus 5} & \parbox[c][3em][c]{4.1em}{\centering GPT-6\\Astra} & \parbox[c][3em][c]{4.1em}{\centering GPT-6.1\\Sol} & \parbox[c][3em][c]{4.5em}{\centering Claude\\Opus 5.5} \\
\midrule
\textbf{\textit{Agentic Coding}} & & & & & & & & & \\
Terminal-Bench 2.1 & 87.9 & 84.3 & 87.6 & 89.6 & 86.2 & \textbf{90.2} & 87.4 & -- & 87.6 \\
DeepSWE & 67.9 & 63.7 & \textbf{73.9} & 67.0 & 70.6 & 72.6 & 74.1 & 75.2 & 74.2 \\
NL2Repo-Bench & 57.2 & 53.0 & 62.7 & 63.2 & 60.7 & \textbf{79.9} & -- & -- & -- \\
SWE-bench Pro & 60.6 & 60.3 & 61.8 & 65.3 & 59.8 & \textbf{76.5} & -- & -- & 89.9 \\
\midrule
\textbf{\textit{Productivity}} & & & & & & & & & \\
SpreadsheetBench 2 & \textbf{37.1} & 32.4 & 35.8 & 34.9 & 32.7 & 33.6 & -- & 32.0 & 35.1 \\
ALE-CLI & 32.4 & 28.1 & \textbf{35.4} & 24.8 & 28.6 & 29.5 & 33.3 & 29.5 & 34.3 \\
APEX-Agents & 61.7 & 50.8 & \textbf{69.2} & 66.3 & 55.4 & 61.3 & 64.7 & 60.0 & 73.5 \\
GDPval & 78.0 & 67.7 & 77.6 & 79.6 & 69.6 & \textbf{80.0} & -- & -- & -- \\
Workspace-Bench-Lite & 67.1 & 65.1 & 66.8 & 67.4 & 57.4 & \textbf{68.1} & -- & -- & -- \\
Workspace-Bench & 61.7 & 59.7 & 63.0 & \textbf{63.9} & 50.9 & 62.0 & -- & -- & -- \\
\midrule
\textbf{\textit{Tool Use}} & & & & & & & & & \\
OfficeQA Pro & \textbf{72.2} & 64.7 & 69.7 & 66.2 & 66.4 & 67.2 & -- & -- & -- \\
Harvey LAB & \textbf{25.0} & 20.0 & 11.7 & 24.2 & 5.0 & 22.7 & -- & -- & -- \\
AutomationBench & 56.7 & 48.3 & \textbf{57.8} & 49.7 & 45.7 & 49.4 & 41.4 & 49.3 & 58.0 \\
BFCL V4 & 75.5 & 71.7 & 71.4 & 66.8 & 68.8 & \textbf{80.6} & -- & -- & -- \\
JobBench & 58.8 & 56.5 & 58.5 & 52.7 & 45.4 & \textbf{68.0} & 50.4 & -- & -- \\
SkillsBench & 62.3 & 58.3 & 52.8 & \textbf{66.7} & 62.5 & 63.7 & -- & 72.6 & 71.8 \\
$\tau^3$-bench Banking & 45.4 & 39.2 & 47.4 & \textbf{55.2} & 46.9 & 48.7 & -- & -- & -- \\
\midrule
\textbf{\textit{ML Engineering}} & & & & & & & & & \\
MLE-bench Lite & \textbf{90.3} & 84.3 & 89.2 & 82.2 & 80.7 & 88.0 & -- & -- & -- \\
PostTrainBench & \textbf{39.2} & 36.8 & -- & -- & 31.0 & 35.7 & 41.9 & 38.8 & 43.8 \\
PaperBench & 88.0 & 83.0 & 88.6 & \textbf{95.1} & 94.3 & 86.9 & -- & -- & -- \\
MLS-Bench & 42.0 & 39.9 & 44.3 & 41.0 & 46.2 & \textbf{49.8} & 50.0 & -- & -- \\
\midrule
\textbf{\textit{Deep Research}} & & & & & & & & & \\
DeepResearch Bench II & 54.2 & 51.9 & \textbf{55.1} & 49.2 & 50.7 & 54.1 & -- & -- & -- \\
\midrule
\textbf{\textit{Search}} & & & & & & & & & \\
DeepSearchQA & 94.4 & 92.6 & \textbf{95.3} & -- & 93.2 & -- & -- & -- & -- \\
BrowseComp-VL & 60.5 & \textbf{61.0} & 60.6 & -- & -- & -- & -- & -- & -- \\
WideSearch & 81.5 & 77.2 & \textbf{86.3} & 81.9 & 83.3 & 84.0 & -- & -- & -- \\
BrowseComp & 91.2 & 88.3 & \textbf{93.8} & -- & 92.2 & 90.8 & 91.5 & 87.5 & 90.8 \\
\midrule
\textbf{\textit{Vision}} & & & & & & & & & \\
OSWorld v2.1 & 69.3 & 66.4 & 70.1 & 66.4 & 70.4 & \textbf{74.0} & 75.8 & 75.2 & 81.8 \\
OSWorld-Verified & 85.2 & 83.5 & 78.5 & 84.3 & 82.2 & \textbf{86.4} & -- & -- & -- \\
WebArena-Verified & 67.6 & 65.0 & 60.7 & 64.9 & 66.6 & \textbf{69.1} & -- & -- & -- \\
Vision2Web & 78.0 & 75.8 & -- & 75.2 & 73.5 & \textbf{80.2} & -- & -- & -- \\
WeaveBench & 68.4 & 60.9 & -- & 71.9 & 79.6 & \textbf{86.4} & -- & -- & -- \\
OSWorld-G & 84.8 & 83.7 & 61.4 & \textbf{88.6} & 77.7 & 76.8 & -- & -- & -- \\
BabyVision & 75.8 & 51.3 & 58.0 & \textbf{83.8} & 80.7 & 63.1 & -- & -- & -- \\
\bottomrule
\end{tabular}}
\end{table}

\subsection{Main Results}

The results show that \model{} supports scientific workflows involving experiments, evidence gathering, and software interaction. Its principal strengths lie in ML engineering, professional deliverables, tool orchestration, and visual interaction. We organize the results around these connected capabilities.

\paragraph{ML engineering and research implementation.}
\model{} achieves 90.3 HumanRank on MLE-bench Lite, ranking first against the five primary baselines. Its PostTrainBench score of 39.2 also leads the primary baselines with available results. These evaluations cover ML problem solving and model post-training, respectively, requiring agents to translate objectives into executable solutions and assess them through experimental feedback~\citep{chan2025mlebench,rank2026posttrainbench}. The results show that the model's coding capabilities support practical ML work, including constructing solutions, executing training procedures, and checking experimental outputs.

The score of 88.0 on PaperBench Code-Dev provides complementary evidence of research implementation. This evaluation covers all 20 papers and requires agents to translate research descriptions into code that can be assessed~\citep{starace2025paperbench}. Together with MLE-bench Lite and PostTrainBench, it examines several stages of research work, from engineering solutions and model training to implementing research methods.

These research workflows also depend on effective use of the computing environment. A score of 87.9 on Terminal-Bench~2.1 demonstrates terminal workflow execution, complementing the ML evaluations with tasks involving data processing, program execution, and result inspection. Scientific objectives ultimately have to be realized through code, experiments, and verifiable outputs; these benchmarks examine key parts of that process.

\paragraph{Tool orchestration for professional deliverables.}
\model{} achieves the highest scores among the five primary baselines on OfficeQA Pro, Harvey LAB, and SpreadsheetBench~2, and ranks second on AutomationBench, BFCL V4, and JobBench. These results span the process of obtaining information, operating tools, and producing a deliverable.

OfficeQA Pro's score of 72.2 reflects the ability to extract and use evidence from supplied documents~\citep{opsahl-ong2026officeqa}. The leading SpreadsheetBench~2 result extends the assessment to executable, verifiable workbooks. Harvey LAB and AutomationBench require all rubric criteria or state assertions to pass, testing whether an agent maintains multiple requirements throughout execution and satisfies them in the final result.

Scores of 78.0 on GDPval and 67.1 on Workspace-Bench-Lite approach the leading primary-baseline results of 80.0 and 68.1, respectively; the model also ranks second on ALE-CLI. These evaluations show tool capabilities applied to complete professional workflows, beyond individual API calls. They align with our training design, which constructs tasks through cross-environment tool composition and verifies completion through environment states and actual deliverables.

\paragraph{Turning visual information into effective action.}
\model{} scores 85.2 on OSWorld-Verified and 67.6 on WebArena-Verified, ranking second on both. These tasks require repeated interpretation of interface states, action selection, and use of post-action feedback to advance the objective. They therefore assess the organization of successive operations into a complete workflow, in addition to recognition and clicking.

OSWorld-G and BabyVision complement these results with evaluations of target localization and visual reasoning. The second-place score of 78.0 on Vision2Web demonstrates another use of visual information: translating visual requirements into an executable website. Together, these tasks cover interpreting visual inputs, locating interaction targets, constructing implementations, and checking outcomes.

This capability profile matches the progression in our visual curriculum from local perception and grounding to interaction in changing environments. Scientific software requires an agent to understand both the current display and how each action changes the environment and advances the task. Desktop, browser, and visual-development tasks provide complementary views of this ability.

\paragraph{Connecting evidence acquisition with research execution.}
\model{} scores 94.4 on DeepSearchQA and 91.2 on BrowseComp, and ranks second on DeepResearch Bench~II at 54.2. These evaluations extend from targeted retrieval to longer research workflows, covering information discovery, verification of leads, and synthesis across multiple searches.

Information acquisition also supports the other capabilities. Professional deliverables require evidence, ML experiments require technical materials, and visual research tasks may require discovering and interpreting images~\citep{geng2025webwatcher}. BrowseComp-VL's score of 60.5 adds evidence of multimodal retrieval, extending the evaluation from understanding supplied materials to actively acquiring external information. Alongside engineering execution and visual interaction, these results establish several of the capabilities needed to participate in scientific workflows.

\subsection{Trajectory Analysis}
\label{sec:trajectory-analysis}
\label{sec:trajectory-audit}

The preceding evaluations assess the ability to conduct experiments, implement research methods, and complete tool workflows. To examine how these capabilities appear during execution, we compare 48 paired trajectories from \model{} and GLM-5.3-Flash, hereafter the baseline: 17 pairs from Terminal-Bench~2.1, 16 from SWE-bench Pro, and 15 from AutomationBench. We examine tool outputs, code changes, and final artifacts to identify differences in key decisions and their associated outcomes. We use these pairs for qualitative comparison, accounting for differences in execution settings and the evidence available for each case.

\paragraph{Iterative coordination and analysis.}
The coordinator verifies run provenance, pairs trajectories, schedules analysis, and retrieves records requested by the analyzer. Starting from the task requirements and trajectory index, the analyzer identifies a behavioral difference to explain and requests the relevant tool calls, code changes, execution feedback, and artifacts. Evidence retrieval alternates with explanation: new records may support the current account or expose gaps that require a revised question or further inspection.

This interaction deepens the analysis within a case. After identifying different implementations, the analyzer checks whether the difference enters the executed path, changes the final output, and corresponds to the test outcome. Truncated observations require retrieval of their continuation before information is treated as absent. The analysis thus progresses from describing actions to examining the relationship between implementation, output, and grading.

A separate analyzer invocation reviews each case against the same records, checking supporting evidence and alternative explanations. Cycle synthesis combines these reviews, findings, and unresolved questions to propose the next round's case focus, evidence requests, and review criteria. The coordinator schedules subsequent analyses and retrieves the requested evidence; proposed review criteria are assessed before adoption. This cycle broadens coverage across cases and deepens the examination of how implementation choices affect execution outcomes. Across the reviewed cases, we identify the following patterns.

\begin{figure}[htbp]
\centering
\includegraphics[width=0.85\linewidth]{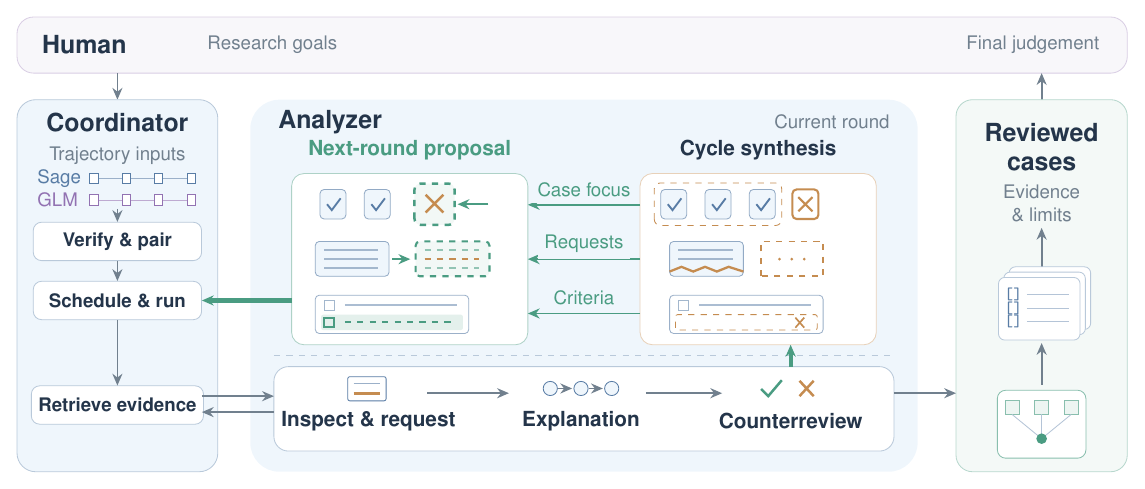}
\caption{Human-guided iterative trajectory analysis. Left: the coordinator verifies and pairs trajectories, schedules analysis, and retrieves requested tool calls, code changes, execution feedback, and artifacts. Lower center: the analyzer inspects records, requests evidence, and develops an explanation, which a separate invocation counterreviews; bidirectional arrows represent on-demand retrieval. Upper center: cycle synthesis proposes the next round's case focus, evidence requests, and review criteria. Dashed elements denote proposals for subsequent analysis. Right: case findings retain evidence and interpretive limits for human review. The cycle supports both broader coverage of cases and questions and closer inspection of the evidence behind each explanation.}
\label{fig:trajectory-analysis}
\end{figure}

\paragraph{Resolving key uncertainties through experimental checks.}
In scientific and computational tasks, \model{} demonstrates the ability to turn implementation hypotheses into executable checks. In the DNA primer design task, both models completed the main sequence design and melting-temperature calculations. \model{} additionally examined ambiguities in matching at the junction, revised the design, and passed the evaluation; the baseline's artifact failed the temperature-difference check. In the ELF extraction task, \model{} tried different compilation configurations and ultimately used the original virtual addresses, whereas the baseline added an extra base address and failed the reference comparison. These trajectories illustrate the use of experimental checks to examine competing interpretations and guide implementation, a process relevant to scientific problem solving.

\paragraph{Preserving scientific correctness through artifact generation.}
In several cases, both models completed similar domain analyses, but their outcomes diverged during artifact generation. In the Bayesian network task, both recovered the same graph structure. The baseline reversed parent and child directions when exporting the intervened graph, and its checks did not cover the semantics of the output columns. \model{} exported the graph in the required direction. In the protein assembly task, both completed the main domain subproblems. The baseline removed a tag from the source sequence according to domain convention and validated against a sequence processed with the same assumption, leaving the conflict with the task requirements undetected. \model{} retained the required full sequence and passed verification. These comparisons show that completing scientific tasks requires consistency among source data, permitted transformations, and output representations. In these cases, \model{} carried correct intermediate analyses into artifacts that satisfied the requirements.

\paragraph{Implementing new requirements while preserving compatibility.}
Software engineering cases reveal how the models handle the relationship between new requirements and existing implementations. In the NodeBB email-confirmation task, the baseline retained the old route and passed local tests without implementing the required new flow; \model{} updated the target route and the associated registration-state handling. In the ProtonMail modal task, the baseline used an opacity setting to keep an existing test passing, whereas \model{} implemented the required visible behavior. Interface evolution exposed a related difference: in the Tutanota draft-folder task, the baseline added a parallel interface while retaining the old one. \model{} supported the new context through the existing interface while preserving old call forms, and checked folder hierarchies and boundary conditions. In the Vuls version-identification task, the baseline retained an outdated branch affecting some inputs, while \model{} revised that branch and checked multiple classes of version strings. These behaviors reflect joint consideration of requirements, calling conventions, and boundary conditions, connecting local code changes to the behavior needed by the system.

\paragraph{Maintaining task constraints across tool operations.}
Differences in tool workflows often concern how information already obtained shapes subsequent actions. In the advertising-pause task, both models identified the main threshold. \model{} additionally read the exception notes and retained the exempt advertisement, whereas the baseline paused it. In the sales-reply task, both computed the same monetary total. \model{} placed the total in the operator-facing report and retained the customer draft's template content; the baseline inserted the amount into the customer draft and triggered the corresponding check. In the ticket-migration task, \model{} preserved the requested priority values, Critical, Major, and Minor, while the baseline mapped them to Highest, High, and Low. These differences arose after the main information had been obtained and concerned its scope, meaning, and intended recipient. In these cases, \model{} maintained a closer correspondence between task rules and the operations performed.

\paragraph{Completing long workflows with verifiable deliverables.}
In the Git web-server task, \model{} checked cloning, pushing, and page access using the specified user and access method. The baseline used a substitute user, and the evaluator subsequently reported an HTTP 404. In the video move-extraction task, \model{} completed the processing workflow and delivered the result, whereas the baseline timed out without producing the final file. Cases in which both succeeded, including database recovery, rendering, and data resharding, show that complex tool operation and error recovery are shared capabilities. The clearer differences lie in whether verification covers the key requirements and whether the final deliverable satisfies them. Inconsistent counts in \model{}'s HelpScout weekly report show that successful individual operations still require a final check for agreement among detailed records, aggregate counts, and completion claims. Together, the successful and unsuccessful cases connect the execution capabilities assessed above to observable experimental checks, constraint preservation, interface implementation, and final delivery.

\section{Application}
\label{sec:application}
\providecommand{\applicationpagebreak}{}

Scientific research requires repeated interaction with heterogeneous data, code, and specialized software. \model{} connects scientific multimodal understanding with tool use to carry out these workflows, from organizing imaging data and constructing three-dimensional models to operating analysis software and investigating questions through the literature. Across the demonstrations below, it inspects intermediate outputs, revises its actions in response to visual or computational feedback, and checks results against source data or task constraints. The resulting editable models, measurement records, and interactive explanations make both the outputs and the reasoning process accessible for researchers to inspect and refine.

\subsection{CT Reconstruction}
\label{sec:applications-ct}
\begin{figure}[htbp]
\centering
\begin{subfigure}[t]{0.325\linewidth}\centering
\includegraphics[height=0.625\linewidth]{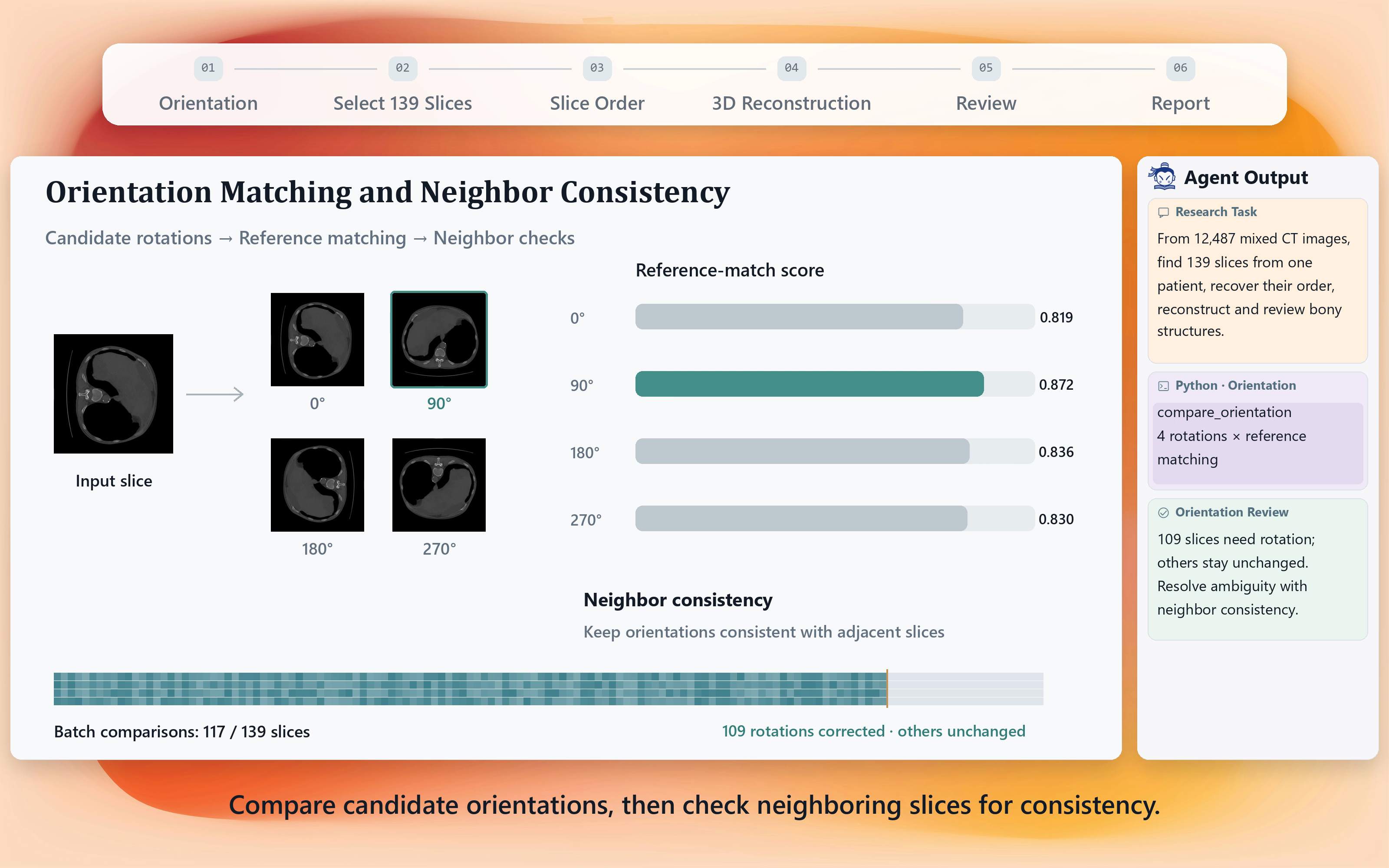}
\caption{Orientation checks}
\end{subfigure}\hfill
\begin{subfigure}[t]{0.325\linewidth}\centering
\includegraphics[height=0.625\linewidth]{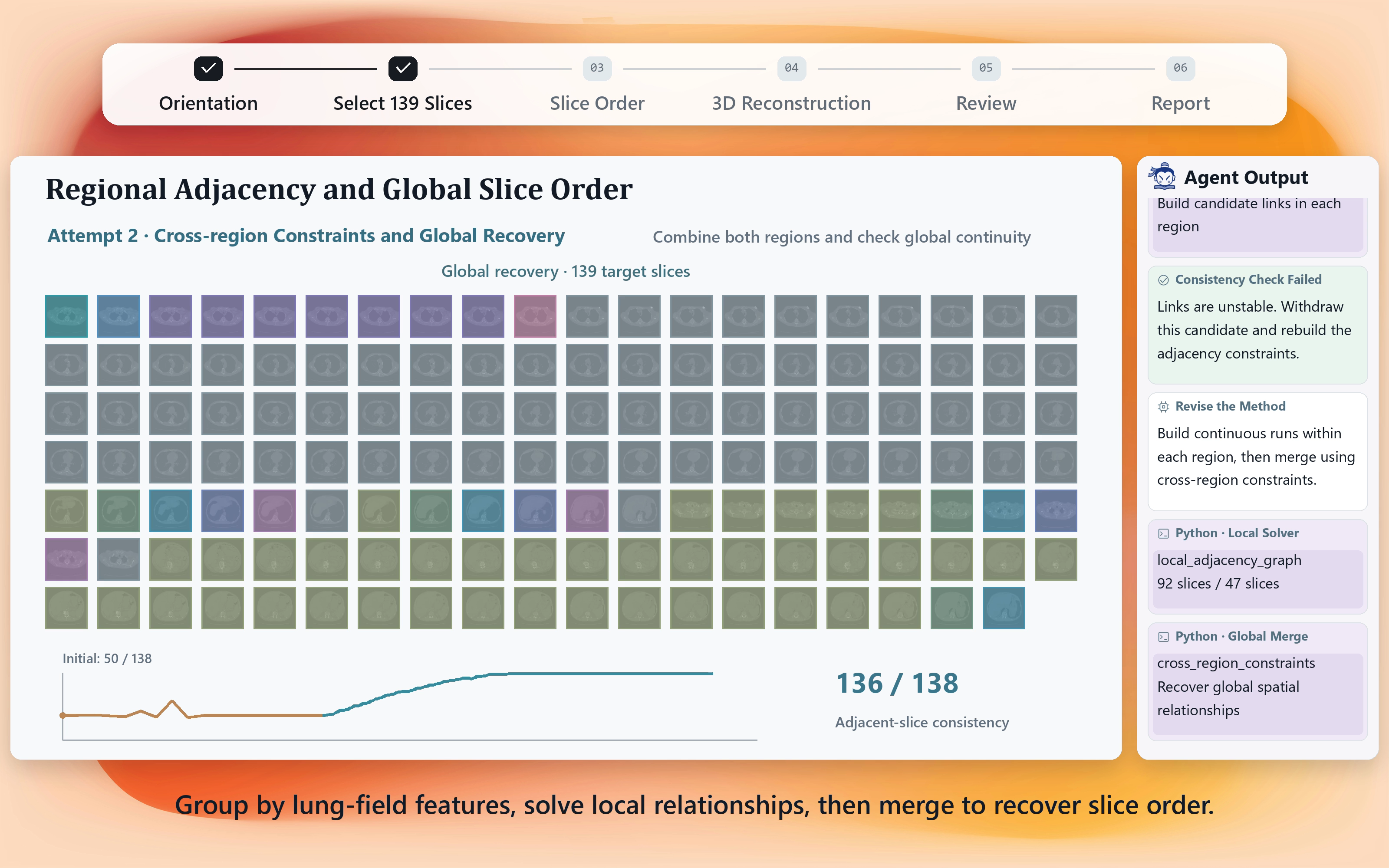}
\caption{Slice-order recovery}
\end{subfigure}\hfill
\begin{subfigure}[t]{0.325\linewidth}\centering
\includegraphics[height=0.625\linewidth]{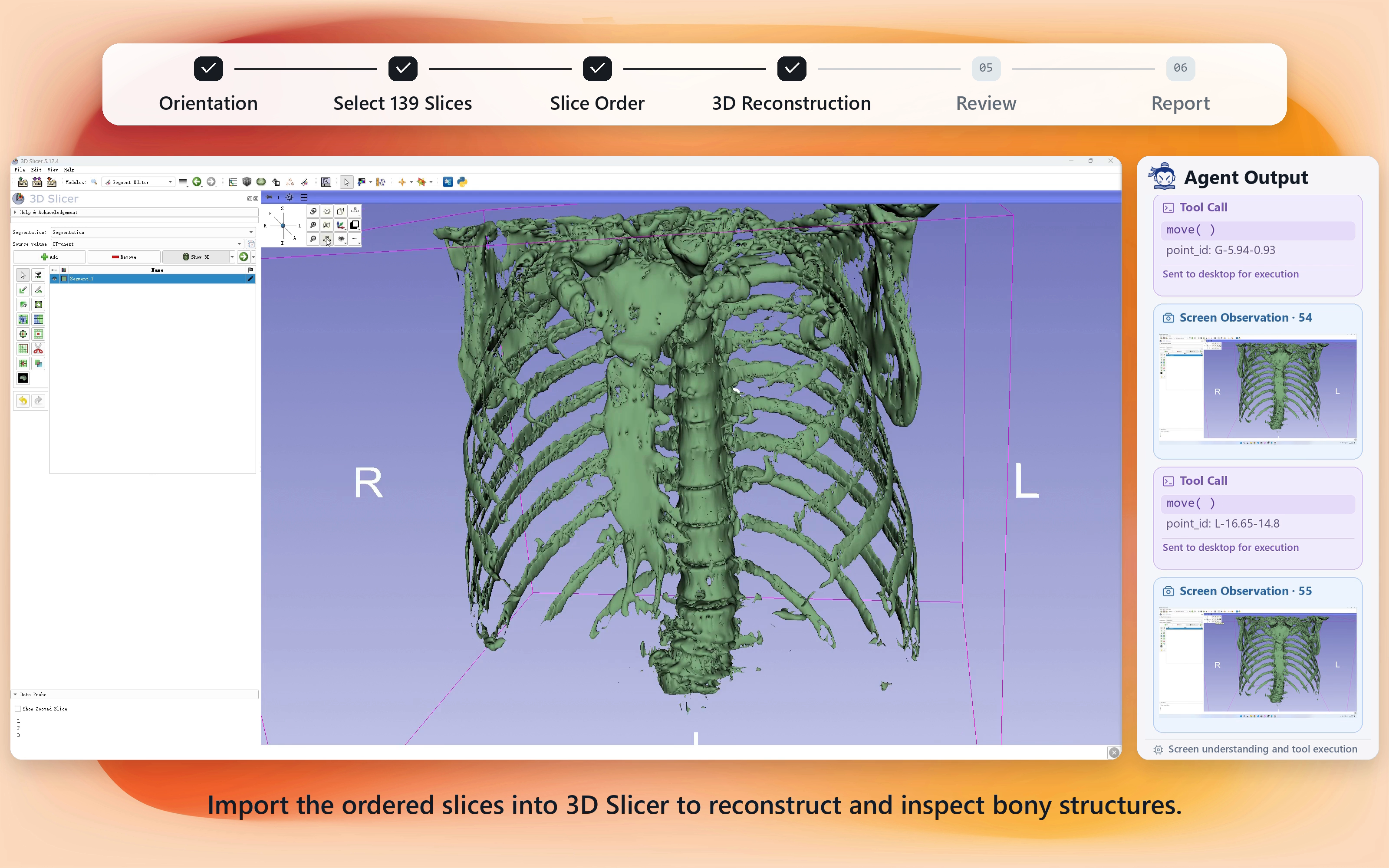}
\caption{3D reconstruction}
\end{subfigure}
\caption{CT reconstruction: orientation matching, sequence recovery, and skeletal visualization in 3D Slicer.}
\label{fig:applications-ct}
\end{figure}

From 12,487 mixed CT images, \model{} identifies 139 slices from a single scan, corrects their orientations, and recovers their order. When the initial sequence fails consistency checks, it reconciles local slice groups into a coherent volume. It then uses 3D Slicer to reconstruct bony structures and returns to the source slices to inspect local anatomical details (Figure~\ref{fig:applications-ct}).

\FloatBarrier

\subsection{Reference-Guided 3D Modeling}
\label{sec:applications-modeling}
\begin{figure}[htbp]
\centering
\begin{subfigure}[t]{0.325\linewidth}\centering
\includegraphics[height=0.51\linewidth]{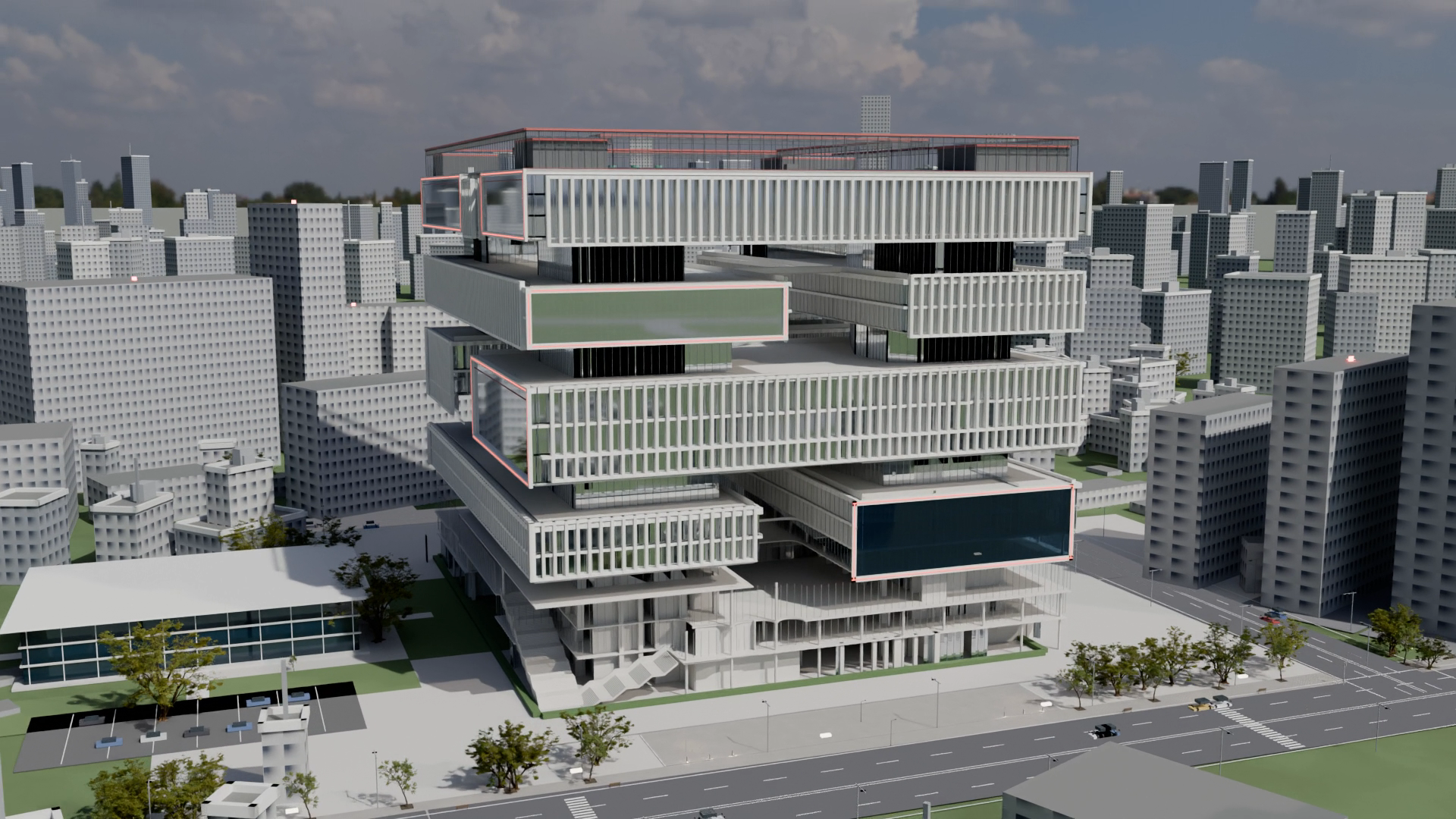}
\caption{Pujiang building}
\end{subfigure}\hfill
\begin{subfigure}[t]{0.325\linewidth}\centering
\includegraphics[height=0.51\linewidth]{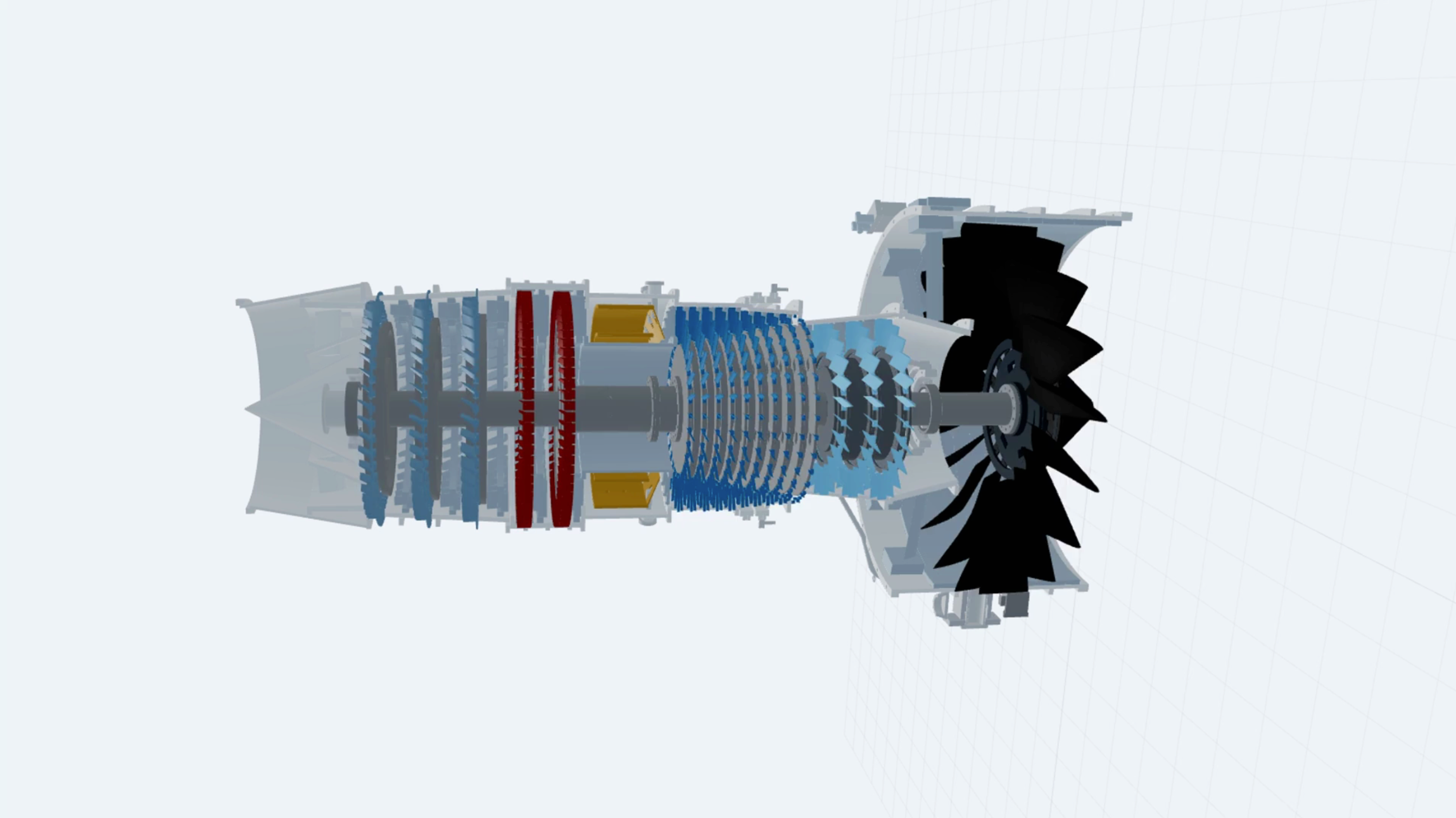}
\caption{Turbofan engine}
\end{subfigure}\hfill
\begin{subfigure}[t]{0.325\linewidth}\centering
\includegraphics[height=0.51\linewidth]{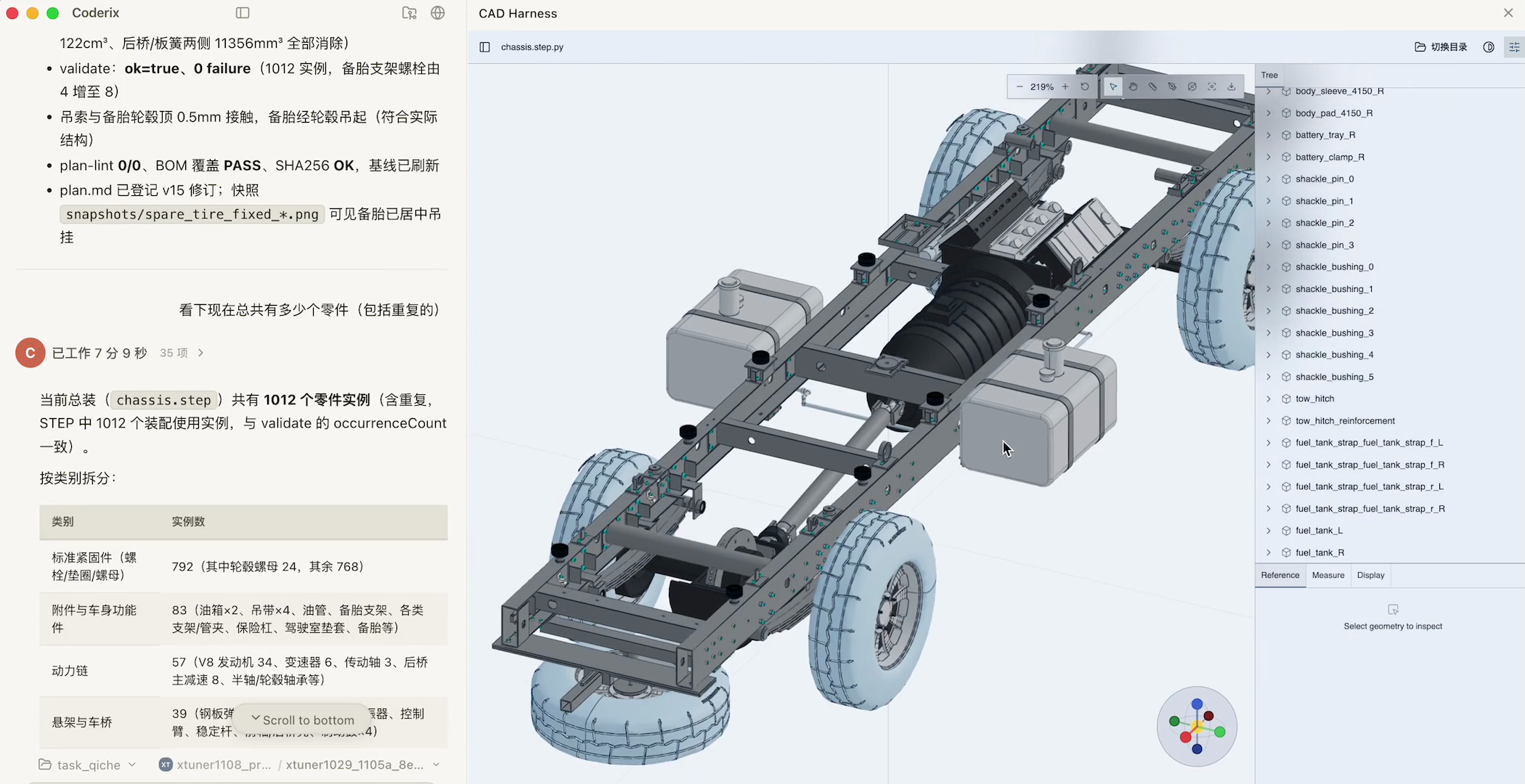}
\caption{Truck chassis}
\end{subfigure}
\caption{Reference-guided architectural and mechanical modeling in Blender and CAD.}
\label{fig:applications-modeling}
\end{figure}

Given reference images, \model{} constructs three editable models (Figure~\ref{fig:applications-modeling}). In Blender, it reconstructs the Pujiang building, refines materials and lighting through rendered previews, and creates a presentation animation. In CAD, it builds a parametric turbofan engine and a truck chassis, coordinating component interfaces and checking geometry and interference. The outputs include editable projects, a STEP assembly, a component inventory, and rendered views for subsequent inspection and design revision.

\FloatBarrier
\applicationpagebreak

\subsection{Scientific Software Analysis}
\label{sec:applications-software}
\begin{figure}[htbp]
\centering
\begin{subfigure}[t]{0.325\linewidth}\centering
\includegraphics[height=0.5625\linewidth]{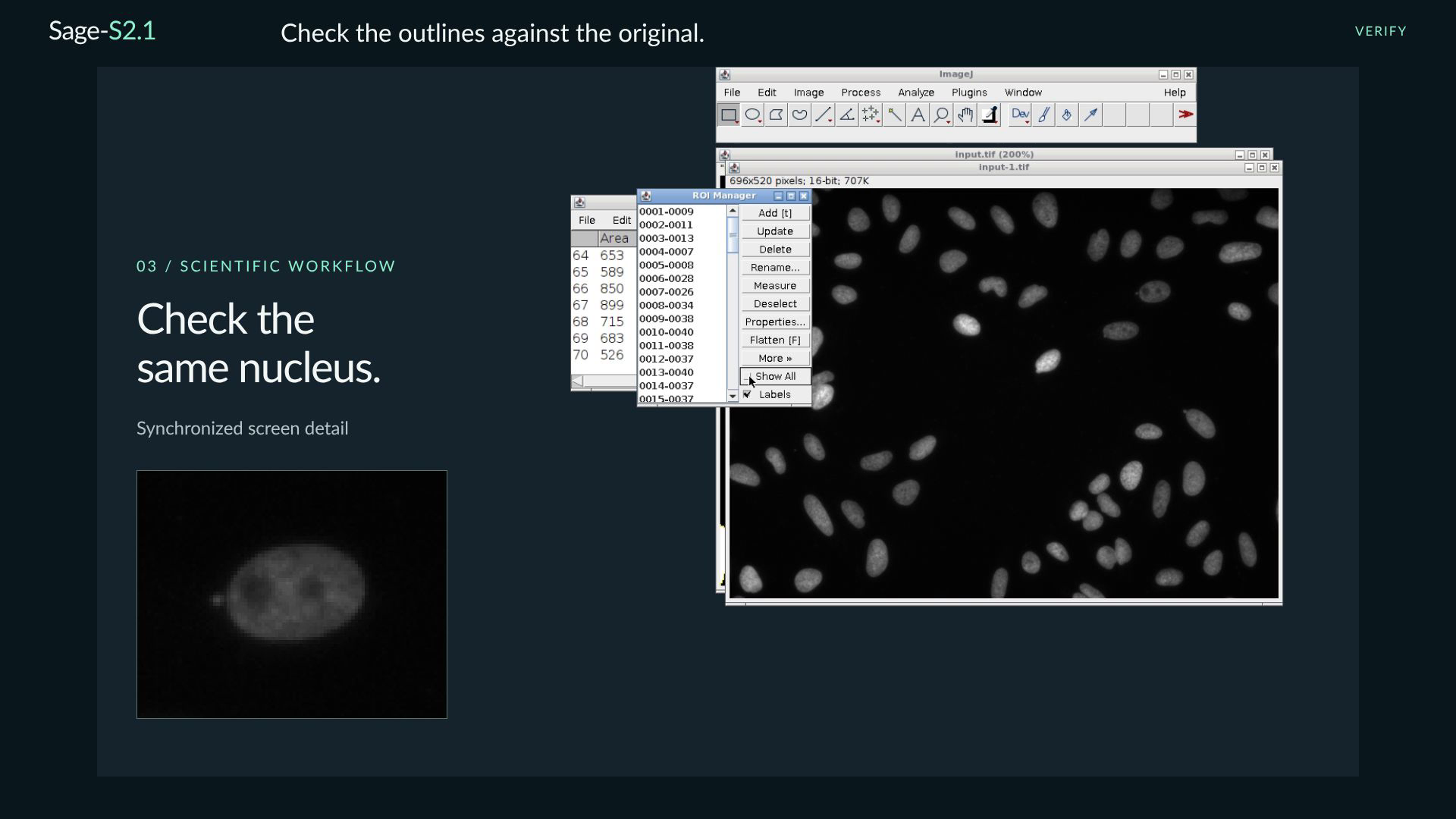}
\caption{Microscopy: Fiji}
\end{subfigure}\hfill
\begin{subfigure}[t]{0.325\linewidth}\centering
\includegraphics[height=0.5625\linewidth]{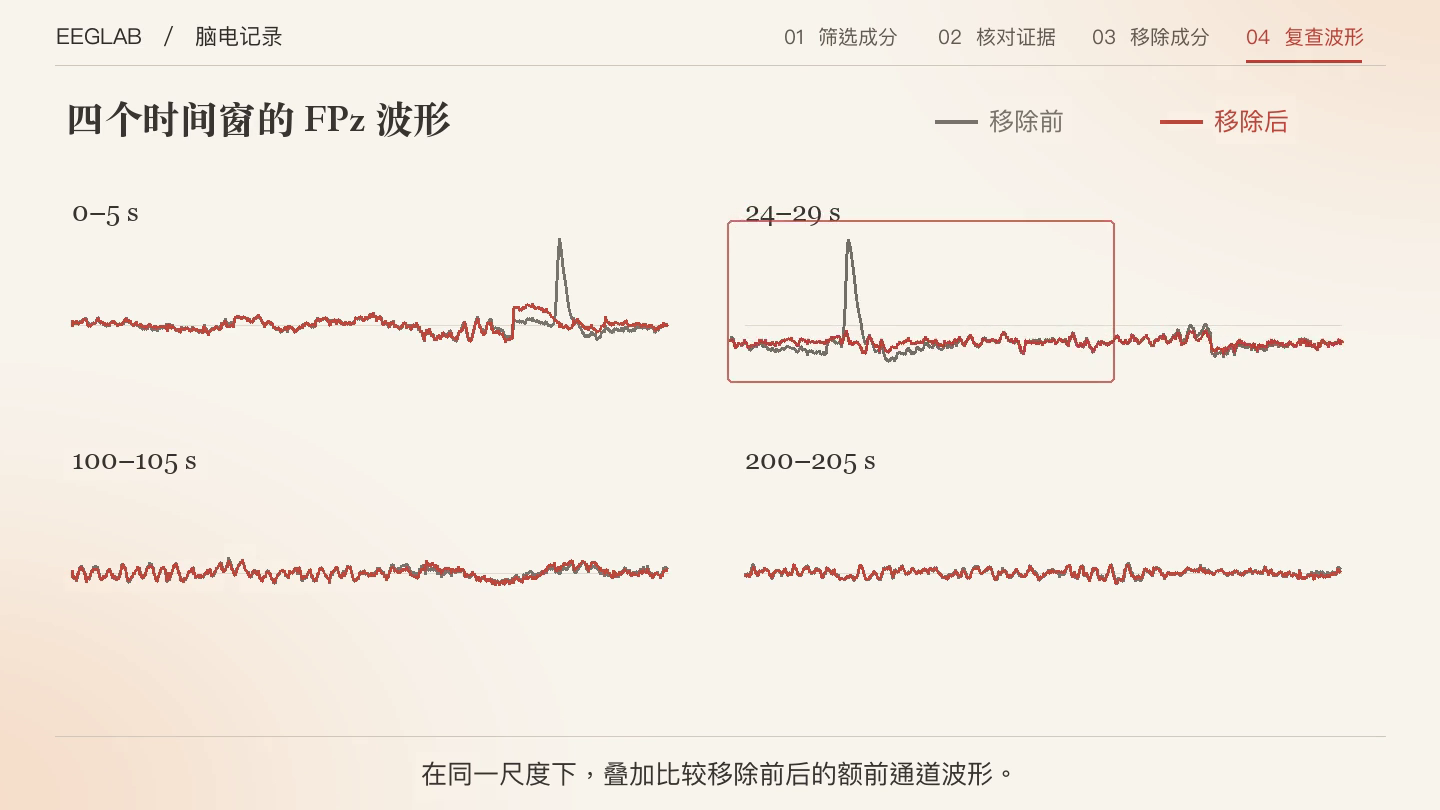}
\caption{EEG: EEGLAB}
\end{subfigure}\hfill
\begin{subfigure}[t]{0.325\linewidth}\centering
\includegraphics[height=0.5625\linewidth]{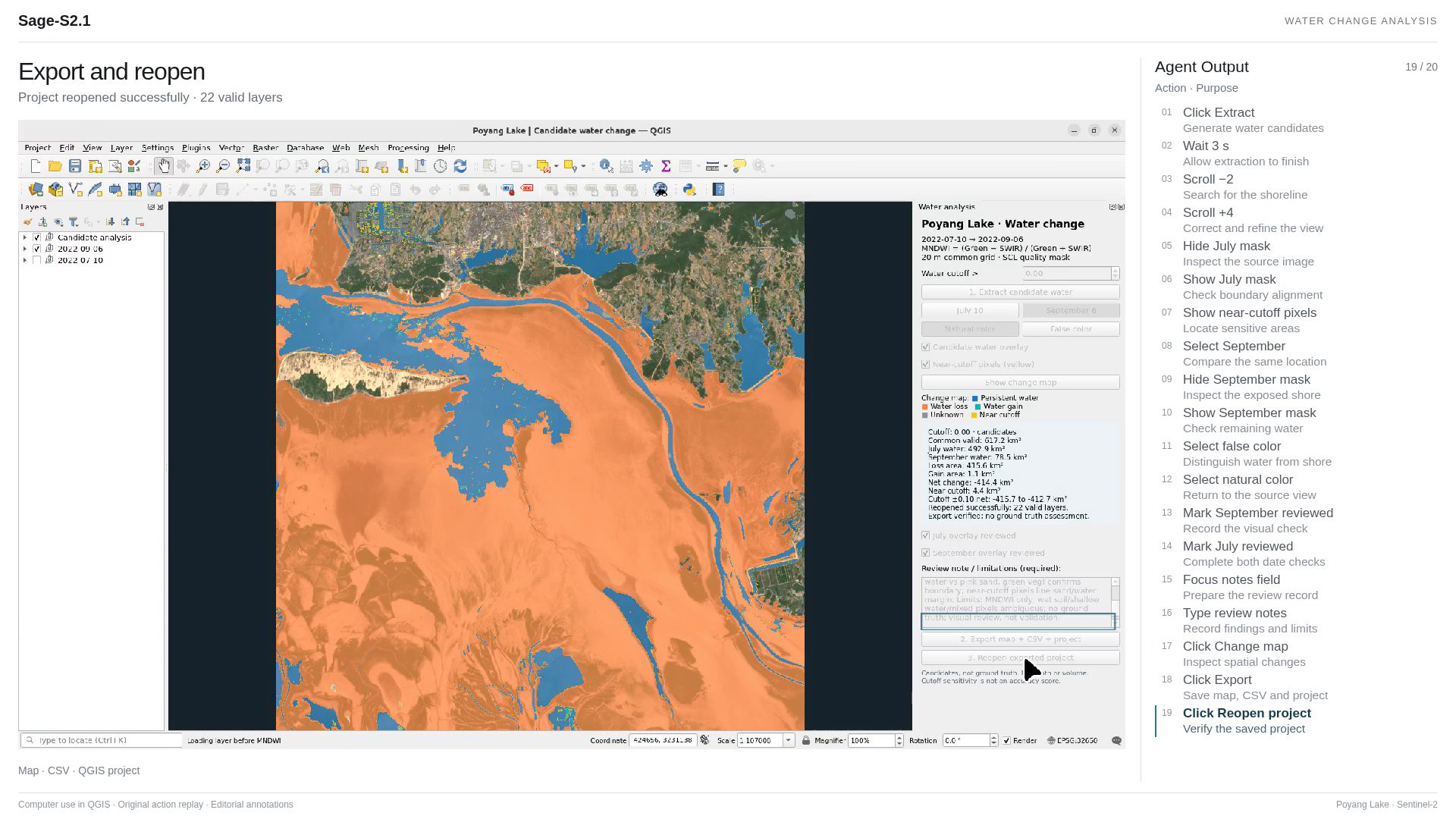}
\caption{Satellite imagery: QGIS}
\end{subfigure}
\caption{Scientific software workflows with visual checks of segmentation, signal processing, and water-area change.}
\label{fig:applications-software}
\end{figure}

\model{} performs three analyses through specialized graphical interfaces (Figure~\ref{fig:applications-software}). In Fiji, it adjusts segmentation thresholds, separates touching nuclei, and overlays contours on microscopy images before exporting regions and measurements. In EEGLAB, it identifies blink-related independent components from temporal, spatial, and spectral features, removes them, and compares waveforms at matching scales across multiple intervals. In QGIS, it extracts and reviews water boundaries from two observations of Poyang Lake, measures changes within their common valid region, and exports a change map, area statistics, and a reopenable project.

\FloatBarrier

\subsection{Scientific Deep Research: The Memory of Light}
\label{sec:applications-memory}
\begin{figure}[htbp]
\centering
\begin{subfigure}[t]{0.325\linewidth}\centering
\includegraphics[height=0.625\linewidth]{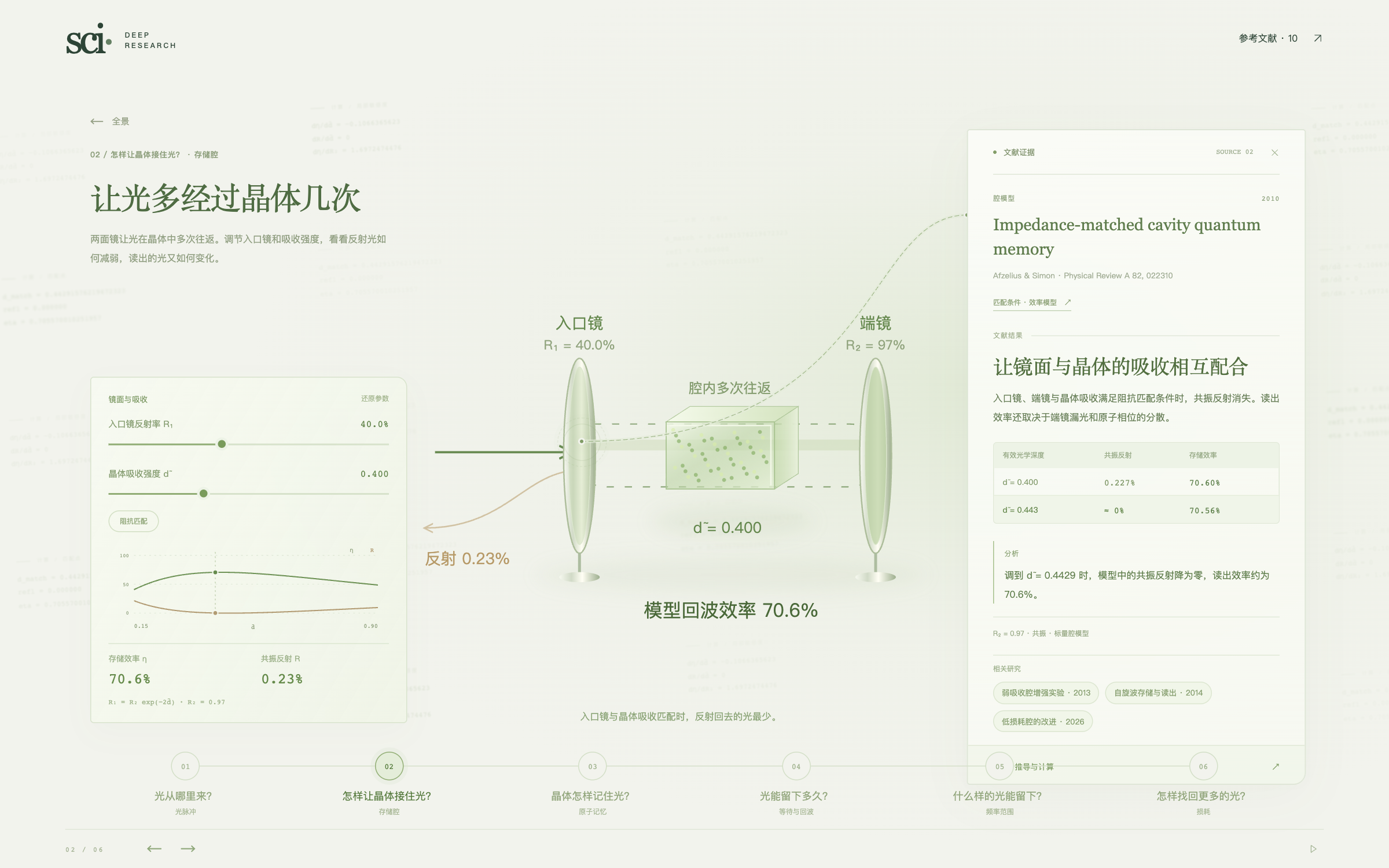}
\caption{Cavity coupling}
\end{subfigure}\hfill
\begin{subfigure}[t]{0.325\linewidth}\centering
\includegraphics[height=0.625\linewidth]{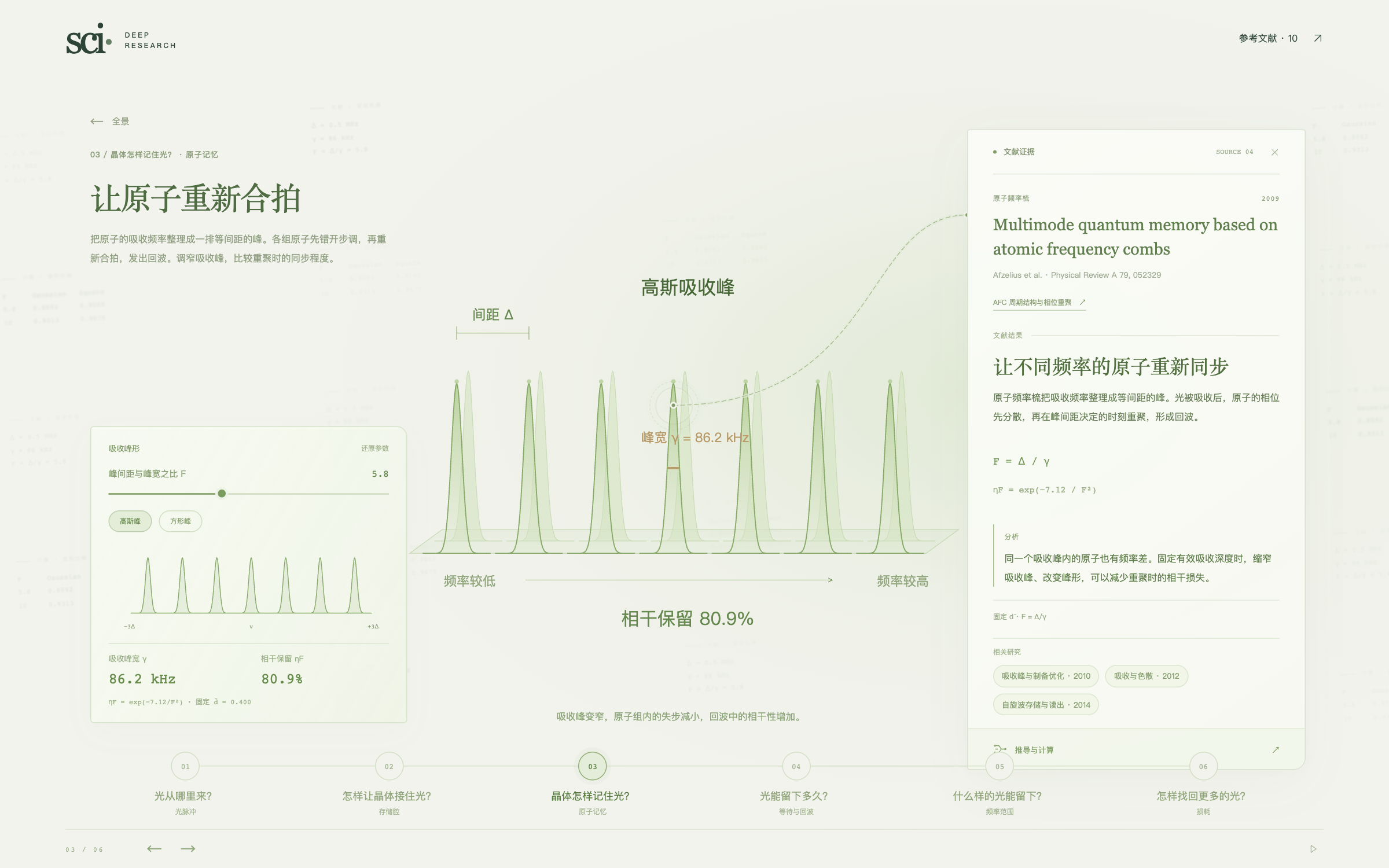}
\caption{Atomic rephasing}
\end{subfigure}\hfill
\begin{subfigure}[t]{0.325\linewidth}\centering
\includegraphics[height=0.625\linewidth]{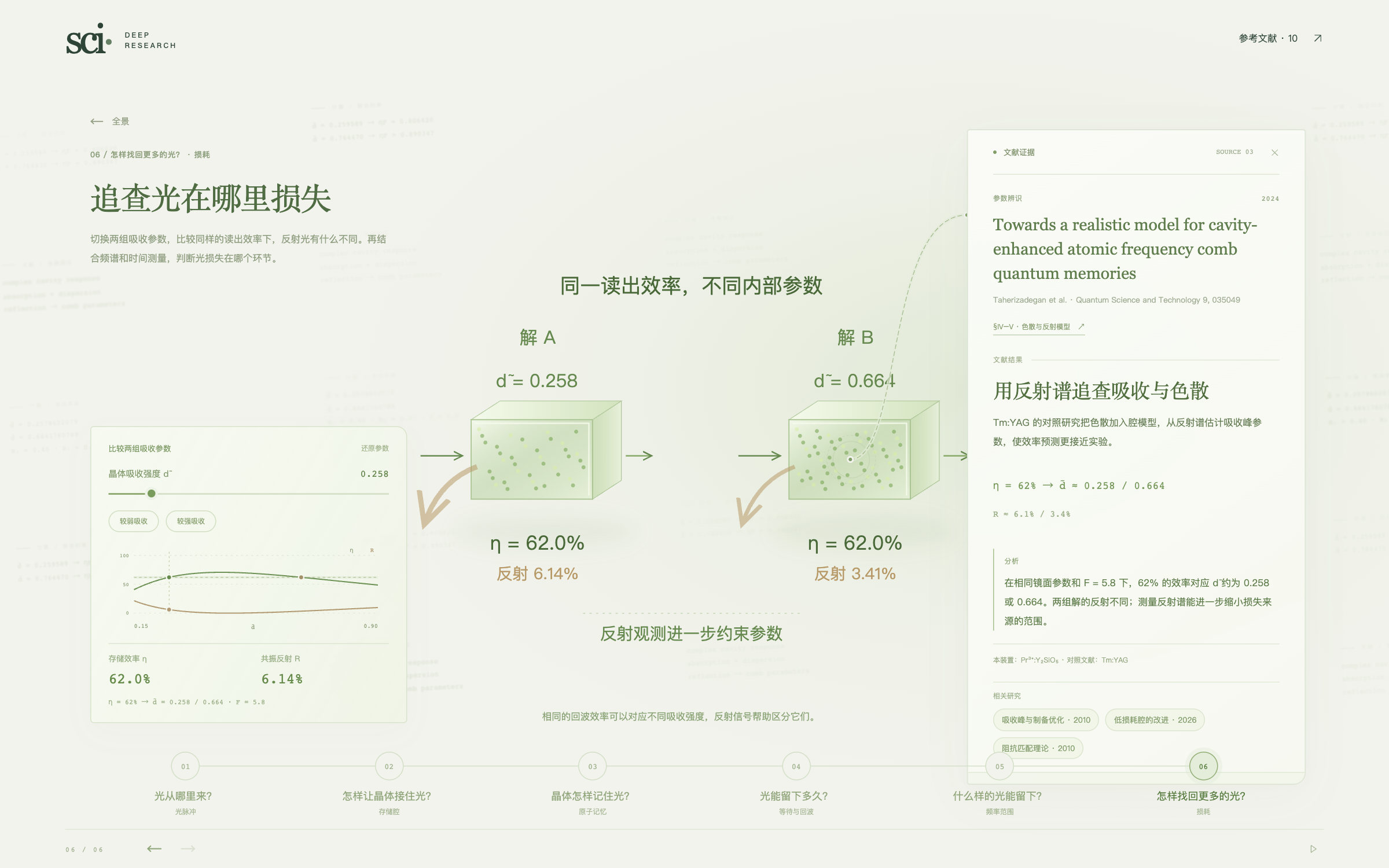}
\caption{Loss analysis}
\end{subfigure}
\caption{The Memory of Light: three interactive views connecting physical parameters, simulated responses, and source literature.}
\label{fig:applications-memory}
\end{figure}

Starting from a reported quantum-memory experiment with 62\% optical-pulse retrieval efficiency, \model{} reviews primary literature, extracts experimental parameters, and compares cavity coupling, atomic dephasing, spectral mismatch, and optical losses through parameter sweeps. It turns the investigation into an interactive HTML application (Figure~\ref{fig:applications-memory}): users vary physical parameters, observe changes in pulses and echoes, and trace explanations to papers and calculations. The loss analysis motivates additional reflection measurements to distinguish competing explanations. Interactive readouts represent the demo's calculations; the 62\% value describes the source experiment.

\FloatBarrier

\section{Conclusion}
\label{sec:conclusion}

\model{} brings native scientific inputs and interactive agent capabilities into a general-purpose model. Scientific modality adaptation trains the projectors to present scientific evidence to the shared backbone, while training in verifiable environments gives the model feedback on whether its actions achieve the task objective.

The evaluation shows particular strengths in knowledge work and ML engineering, including first-place results on OfficeQA Pro and MLE-bench Lite in the primary comparison. \model{} is also competitive on desktop and browser tasks, with room to improve in repository-level coding and web search. The exploratory trajectory analysis illustrates how identical task scores can accompany differences in output verification and reporting. A next step is to test these capabilities together in complete scientific workflows, checking whether the model selects relevant evidence, adapts its actions to feedback, and produces scientifically correct results.

\clearpage
\bibliographystyle{unsrtnat}
\bibliography{refs}

\clearpage
\appendix
\section{Contributions}

The listing of contributors is in alphabetical order based on their last names.

\subsection{Core Contributors}

Kun Cai, Haodong Chen, Guangran Cheng, Qingxiu Cheng, Xu Dai, Guanting Dong, Qiming Ge, Lixin Gu, Tao Gui$^\dagger$, Honglin Guo$^\dagger$, Yuyang Hu, Zixian Huang, Jiajie Jin, Alexander Lam, Jiaqiang Li, Yining Li, Jiahang Lin, Yanjiang Liu, Xinyu Lu, Haijun Lv$^\dagger$, Ningsheng Ma, Zerun Ma, Minlong Peng, Jianmin Qian, Huayu Sha, Junlin Shang, Shuo Shen, Qisheng Su, Jiajun Sun, Yanan Sun, Zhichao Sun, Ran Tao, Zhongbo Tian, Guoqiang Wang, Hui Wang, Rui Wang, Yiding Wang, Zhecan Wang, Hao Xiang, Xinchen Xie, Shuhao Xing, Xiaoyu Xing, Wanghan Xu, Xinyu Yang, Yajie Yang, Dingbo Yuan, Yuhang Zang, Hongjie Zhang, Chengfeng Zhao, Haoran Zhao, Penghao Zhao, Ruojun Zhou, Dongsheng Zhu, Yicheng Zou$^\dagger$

\subsection{Contributors}

Qiye Cai, Yifei Cao, Xinmeng Che, Hongji Chen, Jiabei Chen, Jiadong Chen, Jiahao Chen, Jiayi Chen, Xinjie Chen, Yujia Chen, Xin Deng, Yi Dong, Yichen Dong, Wenjie Fu, Jianing Geng, Chenyang Gu, Ding Han, Han Han, Feiyang Hao, Haotan He, Qiushi He, Xufeng He, Xin Hong, Jie Hou, Binze Hu, Yongkang Hu, Zijian Hu, Enfu Huang, Junhao Huang, Changhao Jiang, Huicheng Jiang, Jiazhen Jiang, Shufan Jiang, Wenqing Jing, Jiahao Kuang, Bowen Lai, Bo Li, Qi Li, Qilong Li, Zhuoqun Li, Tianhao Liang, Jiaxiang Liu, Junzhang Liu, Shaofan Liu, Shuainan Liu, Tong Liu, Yi Liu, Ziyu Liu, Zhonghang Lu, Jianwen Luo, Xuan Luo, Yanyi Luo, Huijie Lv, Mingyu Ma, Houcheng Min, Hongjun Mo, Songhao Ni, Chengjun Pan, Qiyuan Peng, Xiaoxuan Peng, Wanying Ren, Runhui Ruan, Jifei Shan, Zixin Shang, Bing Shao, Zhuohui Sheng, Jiayang Shi, Qingyun Shi, Yang Shu, Aierpanjiang Simayi, Sirui Song, Yuxiao Song, Jiahao Sun, Zhe Sun, Wenzhe Tan, Guoxiang Wang, Hanchen Wang, Jie Wang, Pengbo Wang, Qing Wang, Rui Wang, Weihang Wang, Xiaotao Wang, Yuhui Wang, Zhekai Wang, Zihan Wang, Zile Wang, Jinke Wu, Liwei Wu, Xuansheng Wu, Yida Wu, Yihao Wu, Yinqi Wu, Yue Wu, Yuhan Wu, Yuwa Wu, Zhiheng Xi, Ying Xiao, Zhide Xie, Caijun Xu, Xingle Xu, Yichen Xu, Yongfeng Xu, Xiaolei Yang, Zhixiong Yang, Jiacheng Yao, Dingcheng Yi, Shihong Yi, Jinhui Yin, Peizhou Yu, Shan Yu, Hao Yuan, Junjie Yuan, Xiongyao Zha, Bo Zhang, Caixian Zhang, Kaicheng Zhang, Pengxiang Zhang, Qiuyinzhe Zhang, Wei Zhang, Xiaoqian Zhang, Xinsen Zhang, Jiyuan Zhao, Shuyu Zhao, Pujun Zheng, Zheyu Zheng, Xiaoxue Zhong, Peiheng Zhou, Xiaohao Zhou, Xinyu Zhou, Fangyi Zhu, Guanru Zhu, Yulun Zhu, Jiabao Zhuang

\subsection{Advisors}

Pengfei Cao, Zhicheng Dou, Xianpei Han, Ben He, Guoxiu He, Xuanjing Huang, Tao Ji, Yu-Gang Jiang, Dahua Lin, Hongyu Lin, Kang Liu, Yaojie Lu, Le Sun, Ji-Rong Wen, Jun Zhao, Yutao Zhu

\subsection{Team Leader}

Qi Zhang, Bowen Zhou

\section{Benchmark Evaluation Settings}
\label{app:evaluation-settings}

The configurations below describe evaluations conducted by us. Published scores retain the settings of their source evaluations; the three leading reference columns are excluded from the six-model rankings.

\subsection{Agentic Coding}

\paragraph{Terminal-Bench 2.1~\citep{buchanan2026terminalbench21}.}
Evaluations use Claude Code on 89 tasks. Task timeouts are multiplied by 12, capped at 12 hours; verifier timeouts are multiplied by 3. The verifier uses offline dependencies. Runs allow up to 1,000 interaction steps, a 512K context window, 64K output per generation and max thinking effort. Resources follow each task specification; agents can access network resources through mirrors, while image/PDF reading is disabled. Errors score zero. Each model evaluated here is run in four independent runs and the mean pass rate is reported. The GPT-6 Astra reference score is taken from the official leaderboard~\citep{terminalbench2026leaderboard21}.

\paragraph{DeepSWE~\citep{huang2026deepswe,datacurve2026deepswe11}.}
Evaluations use the AgentCompass framework on DeepSWE v1.1, which contains 113 tasks. The evaluation harness is mini-swe-agent v2.4.2 with the official mini.yaml configuration. Temperature and top\_p are both 1.0. Each task receives 2 CPUs and 8 GiB of memory. Agent and verifier have no external network access; there is no step or cost cap. Commands have a 30-second timeout and model requests a 600-second timeout. The task budget is 6 $\times$ 5,400 seconds, or 32,400 seconds. Hidden tests run in a separate verifier sandbox with a 1,800-second timeout. Pass rate is computed from binary rewards; model timeouts and context overflow count as failures. The reported pass rate is the average of four independent runs per evaluated model.

\paragraph{NL2Repo-Bench~\citep{ding2025nl2repobench}.}
DeepSeek-V4.1-Flash-1017, GLM-5.3-Flash, Claude Opus 5 and \model{} are run with Claude Code, and the other models with OpenHands (v0.56.0). \model{} is evaluated with a 512K context window. The score is the average test pass rate over 104 tasks.

\paragraph{SWE-bench Pro~\citep{deng2025swebenchpro}.}
For SWE-bench Pro, \model{}, GLM-5.3-Flash, DeepSeek-V4.1-Flash-1017 and Claude Opus 5 use Claude Code, while the other evaluated models use OpenHands. The evaluation uses the official V2 locked-protocol prompt wrapper~\citep{dan2026swebenchprov2} and the v1 task set of 731 tasks. Each task receives 4 CPUs and 16 GiB of memory, a 4.5-hour time limit and at most 250 turns. Output is capped at 64K tokens per generation and thinking effort is set to max. External network access is disabled except for model endpoints, and Git history after the base commit is removed to prevent information leakage or benchmark contamination. Image/PDF reading, WebSearch and WebFetch are disabled. Patches are collected using git diff after the agent finishes and graded with official tests in an isolated sandbox.

\subsection{Productivity}

\paragraph{SpreadsheetBench 2~\citep{zhu2026spreadsheetbench2}.}
GPT-5.6 Sol uses Codex, while the other evaluated models use Claude Code. Visualization tasks are graded by Qwen3.7-plus~\citep{qwen2026qwen37plus} as a VLM verifier, while the remaining tasks use rule-based verifiers. The reported score is the weighted average accuracy over 321 tasks. Evaluations use the Harbor framework, with 1 CPU and 10 GB of memory allocated to each task. Each model request has a 60-minute timeout, and there is no interaction-step limit. For Visualization tasks, charts are rendered on Windows before grading.

\paragraph{ALE-CLI~\citep{sun2026agentslastexam}.}
Evaluations use the CLI subset of Agents’ Last Exam v1 with Claude Code. Agents complete multi-step workflows in the prescribed software environments and submit deliverables, checked against version-matched hidden references and deterministic graders. Environments and task assets follow v1. Our configuration differs from the official evaluation setup in the following respects: we use a 512K context window, set max\_tokens to 96K, and allow 10 hours per task. For business\_finance/pe\_screening\_memo\_1/zscaler\_fy2025, grading uses gpt-6.1-sol instead of the gpt-4o-mini specified in the official configuration.

\paragraph{APEX-Agents~\citep{mercor2026apexagents11}.}
Evaluations use v1.1 task definitions and grading rubrics for cross-file workflows in investment banking, management consulting and corporate law. Agents produce deliverables from task materials for rubric-based evaluation. A task succeeds only if all rubric criteria pass. We report pass@1, the task success rate for a single attempt. Execution uses an in-house harness and retains the v1.1 grading changes for noncommittal answers. The interaction-step limit is higher than in the official configuration.

\paragraph{GDPval~\citep{patwardhan2025gdpval}.}
Evaluations use the AgentCompass framework on 220 tasks. \model{} and GLM-5.3-Flash use Claude Code; DeepSeek-V4.1-Flash-1017, Qwen3.8-Max, GPT-5.6 Sol and Claude Opus 5 use OpenHands. Each task receives 4 CPUs and 8 GiB of memory, a 4-hour agent timeout and at most 250 turns. Output is capped at 64K tokens with max thinking effort. WebSearch, WebFetch and image/PDF reading are disabled. GLM-5.2 at temperature 0 judges candidate deliverables against Claude Opus 4.8 deliverables, applying rubric criteria and recording wins, losses and ties. The reported metric is the number of wins divided by the total number of evaluations. Ties remain in the denominator but contribute no wins; no half-win credit is assigned.

\paragraph{Workspace-Bench-Lite~\citep{tang2026workspacebench}.}
Evaluations use the English Lite dataset and the AgentCompass framework. GPT-5.6 Sol uses Codex, while the other evaluated models use Claude Code. Each task receives 2 CPUs and 8 GiB of memory and has a two-hour time limit. The LLM judge is doubao-seed-2-1-lite-260915~\citep{bytedance2026seed21}.

\paragraph{Workspace-Bench~\citep{tang2026workspacebench}.}
Evaluations use the English Full dataset. GPT-5.6 Sol uses Codex, while the other evaluated models use Claude Code. Each task receives 2 CPUs and 8 GiB of memory and has a two-hour time limit. The LLM judge is doubao-seed-2-1-lite-260915.

\subsection{Tool Use}

\paragraph{OfficeQA Pro~\citep{opsahl-ong2026officeqa}.}
The 133 tasks in OfficeQA Pro are evaluated using Claude Code as the harness, with network access enabled. Each model request has a 1,800-second timeout, and each task attempt has a 3,600-second timeout, with up to two retries on timeout and a limit of 200 steps. Answers are extracted and scored using EvalScope’s OfficeQA scoring logic with zero tolerance. Tasks without a valid result are counted as incorrect in the overall score across all 133 tasks.

\paragraph{Harvey LAB~\citep{harvey2026lab}.}
Evaluations use a 120-task subset sampled proportionally from the 1,200 official tasks. GLM-5.2 serves as the LLM judge. The reported score is the proportion of tasks that pass every rubric criterion.

\paragraph{AutomationBench~\citep{shepard2026automationbench}.}
Evaluations use v1.0.6 with an in-house harness and include the null-type handling fix from \href{https://github.com/zapier/AutomationBench/pull/13}{AutomationBench PR~13}. Tasks start from predefined business states and use tools to complete workflows. Official assertion-based rubrics check final states; a task passes only when every assertion passes. The reported metric is task pass rate, excluding the simple domain.

\paragraph{BFCL V4~\citep{patil2025bfcl}.}
Web Search tests use Serper, and samples with API failures are scored as incorrect. Evaluations use the official harness at commit 6ea5797 on 5,217 cases, with 16 concurrent workers. Temperature is set to 0.001. Each request has a 120-second timeout and allows at most 2 attempts. The overall score assigns weights of 10\% to Non-Live, 10\% to Live, 10\% to Irrelevance, 30\% to Multi-Turn and 40\% to Agentic. Within Agentic, Web Search and Memory receive equal weight.

\paragraph{JobBench~\citep{li2026jobbench}.}
Evaluations use the official OpenCode harness on the main split of 65 tasks, with 10 concurrent workers. Each task has a 7,200-second timeout and allows at most 2 attempts. Grok-4.3~\citep{xai2026grok43} serves as the judge, and the final score is the mean of the per-task percentage scores. Tasks with API failures or two failed attempts receive a score of zero.

\paragraph{SkillsBench~\citep{li2026skillsbench}.}
Evaluations use OpenHands on the 79-task subset that excludes multimodal tasks. Results are averaged over three runs.

\paragraph{$\tau^3$-bench Banking~\citep{shi2026tauknowledge}.}
Evaluations use the 97 Banking Knowledge tasks and Tau2 commit a2c0247. Both the user simulator and the judge use GPT-5.2~\citep{openai2025gpt52} with low reasoning effort, and the random seed is 300. The alltools configuration provides BM25 search, dense retrieval using text-embedding-3-large~\citep{openai2024embedding3} with the top 10 results, and a read-only knowledge-base shell. The knowledge base contains 698 documents. Each model request has a 180-second timeout and allows up to 3 retries. Simulations have a 1,800-second timeout, while the outer task timeout is 2,400 seconds. Each task allows at most 200 interaction steps.




\subsection{ML Engineering}

\paragraph{MLE-bench Lite~\citep{chan2025mlebench}.}
Evaluations use the Lite dataset and the Claude Code harness, with a 12-hour time limit per run. The reported metric is HumanRank, calculated following \href{https://arxiv.org/abs/2607.28568}{Frontis-MA1}~\citep{yang2026frontisma1}: the mean percentage of human leaderboard participants surpassed by the selected submission.

\paragraph{PostTrainBench~\citep{rank2026posttrainbench,posttrainbench2026v12}.}
Evaluations use v1.2. Scores for GLM-5.3-Flash, GPT-5.6 Sol and Claude Opus 5 are taken from the official leaderboard at posttrainbench.com, a weighted average over 24 benchmark–base-model combinations (6 benchmarks $\times$ 4 base models). The score for \model{} comes from our own evaluations and uses the same weighted average over the 24 benchmark–base-model combinations.

\paragraph{PaperBench~\citep{starace2025paperbench}.}
Evaluations cover all 20 papers of the official Code-Dev variant. GPT-5.6 Sol uses Codex and other models use Claude Code, with max thinking effort. Each task receives 4 CPUs and 16 GB of memory, no GPU, and a 12-hour limit, with search and page-reading access. GLM-5.3~\citep{zai2026glm53} serves as the judge and scores each rubric leaf; the final score is the mean across 20 papers.

\paragraph{MLS-Bench~\citep{lyu2026mlsbench}.}
Evaluations use Claude Code as the execution harness to propose and implement improvements to machine learning methods within the scope of changes permitted by each official task. Task evaluators test transfer across models, datasets and random seeds. Scores use the official normalized task metric.

\subsection{Deep Research}

\paragraph{DeepResearch Bench II~\citep{li2026deepresearchbenchii,drb22026evaluator}.}
Evaluations use Claude Code. All models use the same search tools, and each task has a four-hour time limit. Each result is the mean of three judge scores. GPT-5.5~\citep{openai2026gpt55}, the officially recommended model, serves as the judge.

\subsection{Search}

\paragraph{DeepSearchQA~\citep{gupta2026deepsearchqa}.}
Evaluations use our in-house ReAct search harness, which provides search, visit and python tools. Runs allow up to 500 interaction steps, with a 256K context window and a 64K output limit per generation.

\paragraph{BrowseComp-VL~\citep{geng2025webwatcher}.}
We use a multimodal ReAct search harness with five tools: text\_search, visual\_search, lookup, summarize, and python\_interpreter. Each task allows up to 100 interaction rounds and 100 tool calls in total, with a 512K-token context window, a 64K-token limit per output, and a three-hour time limit. The harness retains the interaction history and requires a final answer as the context window approaches its limit.

\paragraph{WideSearch~\citep{wong2025widesearch}.}
Evaluations use our in-house ReAct search harness, which provides search, visit and python tools. Runs allow up to 500 interaction steps, with a 256K context window and a 64K output limit per generation. The GPT-5.6 Sol score is calculated over 181 tasks. The Qwen3.8-Max score is taken from its official model card.

\paragraph{BrowseComp~\citep{wei2025browsecomp}.}
Evaluations use our in-house ReAct search harness, which provides search, visit and python tools. Runs allow up to 500 interaction steps, with a 256K context window and a 64K output limit per generation. Context is managed with the discard-all strategy.

\subsection{Vision}

\paragraph{OSWorld v2.1~\citep{xlang2026osworld21}.}
GPT-5.6 Sol is evaluated using Codex 0.167.2, while the other evaluated models use Claude Code. Public internet access is disabled for Offline tasks and enabled for Online tasks. WebSearch and WebFetch are disabled for all tasks. Direct PDF input to the model and external OCR are prohibited. PDF text may be extracted locally, or pages may be rendered as images for reading.

\paragraph{OSWorld-Verified~\citep{xlang2025osworldverified}.}
Historical screenshots are folded in batches of 10 steps using a 20-screenshot window. Folded screenshots are replaced with Observation text generated by the model at the corresponding steps. The first and current screenshots are always retained, along with the complete history of text, reasoning and tool interactions.

\paragraph{WebArena-Verified~\citep{hattami2025webarena}.}
Evaluations use the CUA-Lite framework and absolute pixel coordinates. Historical screenshots are folded in batches of 10 steps using a 20-screenshot window.

\paragraph{Vision2Web~\citep{he2026vision2web}.}
GPT-5.6 Sol uses Codex, while the other evaluated models use Claude Code.

\paragraph{WeaveBench~\citep{li2026weavebench}.}
Evaluations use Codex in an Ubuntu sandbox, combining GUI observation and interaction with CLI and code execution to complete tasks across applications. The official trajectory-aware judge applies rubrics to deliverables, screenshots, files and execution logs, and checks for shortcuts such as fabricated screenshots and hard-coded results. Tasks that violate the anti-cheating rules receive a score of zero.

\paragraph{OSWorld-G~\citep{xie2025osworldg}.}
Evaluations use the refined prompt. The model submits pixel coordinates in the original image through click(x,y,button). When no target is visible, it submits (-1,-1,left).

\paragraph{BabyVision~\citep{chen2026babyvision}.}
Images are upscaled with LANCZOS while preserving their aspect ratio so that the shorter side is at least 1,500 pixels. Images that already meet this requirement retain their original dimensions and are not downscaled.

\subsection{Published Reference Results}
These three columns are not our own runs. They are taken from public sources: the vendors' official reports, the Vals AI leaderboard~\citep{vals2026benchmarks}, and the launch tables of Kimi K3.1 Pro, Hy4 and ERNIE 5.5 Flash Preview~\citep{moonshot2026kimik31pro,tencent2026hy4,baidu2026ernie55flash}. Their evaluation settings differ from the other columns, so they are for reference only and are not counted when the best score in each row is highlighted.

\end{document}